\documentclass[journal, 10pt]{IEEEtran}

\usepackage{times} 
\usepackage{mathptmx} 

\usepackage[bidi=default, english]{babel}
\usepackage[LAE, T1]{fontenc}
\usepackage{graphicx}
\usepackage{titlesec}

\usepackage{listings}
\usepackage{amsmath}
\usepackage{xcolor}

\lstdefinestyle{jsonstyle}{
    basicstyle=\ttfamily\small,
    keywordstyle=\color{blue},
    commentstyle=\color{green},
    stringstyle=\color{red},
    breaklines=true,
    showstringspaces=false,
    backgroundcolor=\color{white},
}

\usepackage{amsmath}
\usepackage{array}
\usepackage{tabularray}
\usepackage{tikz}
\usetikzlibrary{arrows.meta, positioning, shapes, arrows, shapes.geometric}

\usepackage{caption}

\usepackage{float}  
\usepackage{lipsum} 
\usepackage{hyperref}

\begin{document}

\title{Invizo: Arabic Handwritten Document Optical Character Recognition Solution}

\author{Alhossien Waly\textsuperscript{\textbf{*}}, Bassant Tarek\textsuperscript{\textbf{*}}, Ali Feteha\textsuperscript{\textbf{*}}, Rewan Yehia\textsuperscript{\textbf{*}}, Gasser Amr\textsuperscript{\textbf{*}}, Walid Gomaa\textsuperscript{\textbf{*}}, \& Ahmed Fares\textsuperscript{\textbf{*}}\\
\textsuperscript{\textbf{*}}Department of Computer Science and Engineering, Egypt-Japan University of Science and Technology,\\
Alexandria, Egypt
\thanks{}
\thanks{The Implementation of this paper can be found in \url{https://github.com/Hedrax/Invizo-OCR}} 
\thanks{}}

\maketitle

\begin{abstract}
Converting images of Arabic text into plain text is a widely researched topic in academia and industry. However, recognition of Arabic handwritten and printed text presents difficult challenges due to the complex nature of variations of the Arabic script. This work proposes an end-to-end solution for recognizing Arabic handwritten, printed, and Arabic numbers and presents the data in a structured manner. We reached 81.66\% precision, 78.82\% Recall, and 79.07\% F-measure on a Text Detection task that powers the proposed solution. The proposed recognition model incorporates state-of-the-art CNN-based feature extraction, and Transformer-based sequence modeling to accommodate variations in handwriting styles, stroke thicknesses, alignments, and noise conditions. The evaluation of the model suggests its strong performances on both printed and handwritten texts, yielding 0.59\% CER and \& 1.72\%  WER on printed text, and 7.91\% CER and 31.41\% WER on handwritten text.\\
The overall proposed solution has proven to be relied on in real-life OCR tasks. Equipped with both detection and recognition models as well as other Feature Extraction and Matching helping algorithms. With the general purpose implementation, making the solution valid for any given document or receipt that is Arabic handwritten or printed. Thus, it is practical and useful for any given context.
\end{abstract}


%
\IEEEpeerreviewmaketitle

\section{Introduction}
\IEEEPARstart{O}{ptical} Character Recognition (OCR) is the process that converts an image of text into machine-readable text format. In a real-world application example, if you scan a form or a receipt, you cannot use a text editor to edit, search, or count the words in the image file. However, you can use OCR to convert the image into a text document, storing its contents as text data. Computer Science has a long history of advancing solutions to this certain task from old classical statistical methods of converting a binarized image of a character or a digit to a text value through Template Matching (\hyperref[Belongie2002]{Belongie et al.,2002}), K-Nearest Neighbors (KNN) (\hyperref[Zhang2006]{Zhang et al.,2006}), or Support Vector Machines (SVM) ((\hyperref[Bahlmann2002]{Bahlmann et al.,2002})) into the advancement of a full LLM solution that can understand the image of document and feedback information on a given prompt (\hyperref[Kim2022]{Kim et al.,2022}).

Any OCR known method must involve an OCR Engine "Recognition model" capable of converting an incoming image to a valid text value. But the recognition-model solution has been seen to have multiple issues regarding reading paragraphs, text with background noise, low-quality images, and the diversity of fonts. To overcome these problems, the OCR module involved more stages in the solution than the recognition stage alone. The first involved image acquisition and input to ensure the proper format of the input text image for further processing. Then, the Pre-Processing Stage (Binarization, Noise Reduction, Normalization) removes background noise and passes an image with a similar background as the ones passed in the training phase of the Recognition Model. The third stage involved the Text Detection and Segmentation module that is responsible for extracting lines of the paragraph images. Such images have multiple lines of text aligned vertically, and segmenting the training images to simulate a background with unrelated noise to the image such as dots and horizontal line of the papers and receipts that we may find on real-world official documents. Even the Character recognition and classification stage itself has been involved in all its sub-stages. The Feature Extraction substage advanced with more techniques to structure different architectures of the Convolutional Neural Networks. The mapping of these features to valid logits is done via different Recurrent Neural Network architectures or Transformer Encoder architectures. In the end, the advancement of the classification head such as CTC Algorithm or Attention base decoder. The final stage added to the system is the enhancement stage for optimization of the output to get higher accuracy. Some used approaches to enhance the results via a predefined word list, such as the one found in CTC-Word-Beam Search (\hyperref[Scheidl2018]{Scheidl et al.,2018}).

\begin{figure}[!hbtp]
    \centering
    \includegraphics[width=1\linewidth]{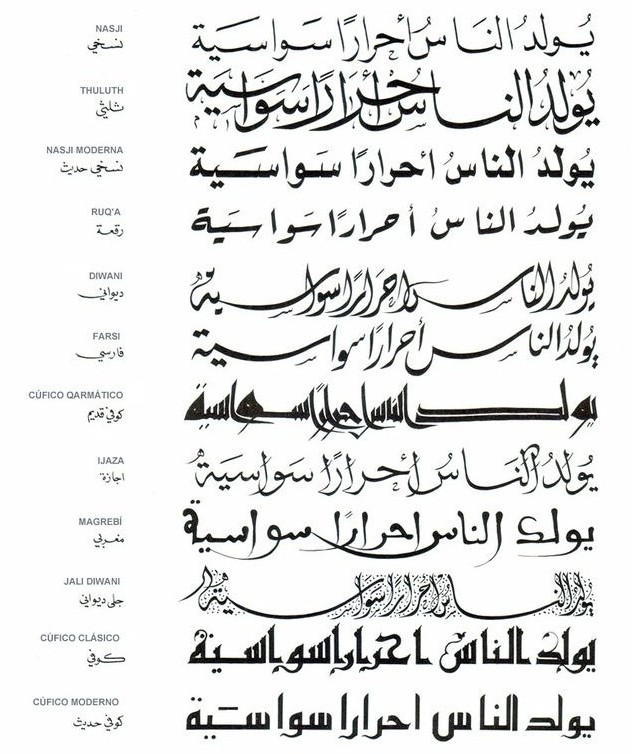} 
    \caption{Some of the different Arabic famous calligraphy fonts}
    \centering
    \label{Arabic calligraphy fonts}
\end{figure}

\begin{figure}[!hbtp]
    \centering
    \includegraphics[width=1\linewidth]{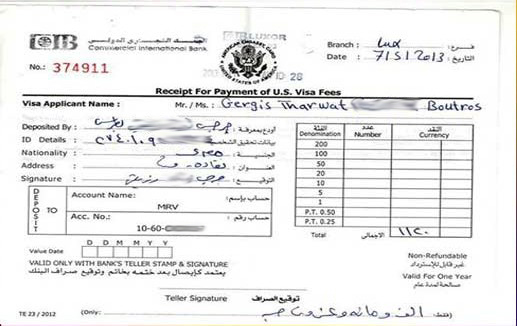} 
    \caption{General example of a document with Complex Structure form}
    \centering
    \label{document with Complex Structure form}
    \vspace{-12pt}
\end{figure}

The previously mentioned problems are general. But when it comes to language-specific OCR recognition problems, we find in Arabic problems due to the Arabic handwritten nature that varies with cursive nature and diverse handwriting styles that can be found among regions and individuals, with cursive forms and artistic calligraphy which makes it even harder to link each styled writing to the Arabic character set. Many Handwritten writing styles have a styling of characters that makes them much closer to another character than the targeted ones and native speakers are only able to identify these characters through the context of the sentence or the closest reasonable word. The use of Diacritics "Tashkeel" (e.g.,\foreignlanguage{arabic}{ َ  ُ  ِ  ْ}) to represent short vowels and pronunciation guides. Many solutions don't include Diacritics so they ignore it in the Ground-Truth labeling of datasets if ever existed in the text images, but Diacritics may form a problem when testing on a real-world image as it might confuse the recognition model to the similarity between it and the regular dot in case of (\foreignlanguage{arabic}{ ُ  ْ}) and a double dot in case of (\foreignlanguage{arabic}{ ِ  َ}).\\
In general Arabic calligraphy has many diverse styles of writing in Arabic as shown in the Figure \ref{Arabic calligraphy fonts}. Those styles are evolved through history and due to the different cultures of Arabic speakers.\\
The complex character connections and ligature as Arabic is a cursive script where letters are connected in most cases. The shape of a letter varies depending on its position in the word (isolated, initial, medial, final). 
For example, the letter "\foreignlanguage{arabic}{ع}" has different forms:\foreignlanguage{arabic}{ع} (isolated),\foreignlanguage{arabic}{عـ} (initial),\foreignlanguage{arabic}{ـعـ} (medial), and \foreignlanguage{arabic}{ـع} (final). Arabic is written from right to left (RTL), which is the opposite of most Western languages that involve left to Right which causes Text Generation change in the Recognition Model. Arabic has letters with visually similar shapes but different meanings, such as\foreignlanguage{arabic}{ ب، ت، ث }which only differ by the number and position of dots (e.g., \foreignlanguage{arabic}{بنت} (girl) and \foreignlanguage{arabic}{بيت} (house)). Also, well-labeled Arabic open-source datasets are not as common as handwritten Latin-character set text. In addition to the datasets that are needed for the recognition task Labeled with Text Ground-Truth, the datasets with Bounding Boxes annotations which are needed for the Text-Detection and Line Segmentation task are also very scarce in the publicly available domain. The datasets problem is a major problem with the advancement of Arabic OCR solutions.
The OCR solutions in real-world applications may come with more complexity than reading a piece of paper. It might involve more complicated processing of more complex structured documents and receipts. Therefore a more complex solution is required to handle the document Segmentation before moving on to the OCR Module (e.g. figure \ref{document with Complex Structure form}).

The Research objective of the OCR solution mentioned has a wide scope. As the solution has many stages, improving each stage will reduce the overall error rate. Enhancing the solution's accuracy can involve optimizing the pre-processing of the image quality, improving Real-Scene 
Text Detection, and enhancing the Character Recognition Rate, all of which fall under the umbrella of Software Architecture. The Arabic Handwritten Solution is a key step towards automating data entry job titles in local and international companies that handle Arabic documents with handwritten parts. If the solution is accurate enough, Arabic data entry job titles will be eliminated, and only one human operator will be required to check the machine's OCR output for validation or further data operations.

In developing an OCR solution capable of handling both complex layouts and Arabic script challenges, we adopted a multi-stage structured methodology that ensures accuracy and efficiency. The methodology includes Synthesized dataset creation, filtering existing datasets with proper labels, pre-processing, model selection, training, post-processing for enhanced results, and several evaluation metrics for each sub-task of the solution. In data handling, Arabic Datasets are very scarce in both recognition and detection tasks. We filtered all labels in a dataset for Arabic handwritten Scene Text Detection to use only the ones with no bad labels and line level annotated.
In the Recognition task, we used all publicly available line handwritten and printed datasets that we could reach in addition to creating a synthesized dataset with 21 different fonts "\foreignlanguage{arabic}{نسخ}-Naskh", "\foreignlanguage{arabic}{رقعه}-Rekaa", "\foreignlanguage{arabic}{ديواني}-Diwany", ..etc. Some of the fonts mimic the basic handwritten shapes. All fonts were carefully picked to generalize the learning over different writings that have a high similarity with human handwritten styles. In the pre-processing, we enhanced model training through augmentation that makes the white background training sample look like the real image with the text after noise filtration. Various experiments done to test different models performance on collected and generated datasets in the process of model selection. The last stage in the solution is the post-processing to optimize the results whenever possible through various solutions including expected labels by the end-user and optimization by Levenshtein Distance (\hyperref[Yujian2007]{Yujian et al.,2007}).

This paper is organized as follows: Section \ref{related work Section} focuses on relevant literature on Optical Character Recognition (OCR) and real-scene text detection, highlighting existing solutions, and challenges unique to Arabic text recognition. Section \ref{methodology Section} describes the proposed approach in detail, including data collection, preprocessing, model architecture, training procedures, and postprocessing techniques used to enhance accuracy. Section \ref{results section} presents the results of the experiments, evaluating the performance of the proposed solution using various metrics, and also discusses the implications of the findings and their relevance to the research objectives. Section \ref{conclusion section} summarizes the study's key findings and highlights its contributions. Section \ref{future work section} identifies potential areas for future research and improvements in OCR systems.
\section{Related Work} \label{related work Section}
\subsection{Text Detection \& Line Segmentation}
Previously the scene text detection process consisted of two stages, the first was to locate the characters or components, second was to group them into words. Neumann and Matas (\hyperref[Neumann2012]{Neumann et al.,2012}) made an end-to-end solution for text localization and recognition by two different stages in the first classification stage to calculate the  probability of each
ER being a character calculated using novel features calculated with O(1) complexity per region tested then choose the ERs with local maximum probability to pass it to the second stage that has improved in classification using more computationally expensive features. A highly efficient exhaustive search with feedback loops is then applied to group ERs into words and to select the most probable character segmentation. Jaderberg (\hyperref[Jaderberg2016]{Jaderberg et al.,2016}) made an end-to-end text spotting pipeline proposed with these stages: a) A combination of region proposal methods extracts many word bounding box proposals. b) Proposals are filtered with a random forest classifier reducing the number of false-positive detections. c) A CNN performs bounding box regression to refine the proposals. d) A CNN performs text recognition on each of the refined proposals. e) Detections are merged based on proximity and recognition results, and a score is assigned. f) Thresholding the detections results in the final text spotting result. Epshtein et al (\hyperref[Epshtein2010]{Epshtein et al.,2010}) uses an enhanced version of the Stroke Width Transform to find letter candidates and group letters into the region of text.\\
Newly, deep learning has dominated the scene text detection area. The deep-learning-based scene text detection methods are classified into many categories according to the granularity of the predicted target The most important of them are: part-based methods, regression-based methods, Sequence-Based methods, and segmentation-based methods.\\
Part-based methods first detect small parts by breaking the text into small parts characters or sub-regions then combine them into word or line-bounding boxes. Shi (\hyperref[Shi2017]{Shi et al.,2017}) introduce Segment Linking (SegLink) the core concept is to break down the text into two locally identifiable components: segments and links. A segment is an oriented box that refers to a part of a word or text line; a link connects two neighboring segments, signifying that they are part of the same word or text line. Both components are detected densely across various scales using an end-to-end trained, fully-convolutional neural network. The final detections are generated by merging segments that are connected by links. Tang (\hyperref[Tang2019]{Tang et al.,2019}) has overcome the difficulty in handling curved and dense texts in (\hyperref[Shi2017]{Shi et al.,2017}) by instance-aware component grouping (ICG) algorithm which is a flexible bottom-up method. To tackle the challenge of distinguishing between dense text instances that many bottom-up methods encounter, we introduce an attractive and repulsive link between text components that encourages the network to concentrate more on nearby text instances, along with instance-aware loss that Optimally uses the context to guide the network effectively. Despite the Part-based method's effectiveness, it requires a lot of complexity and time to link between text components also it may face problems with text variation in size or shape which makes them hard to tune.\\
regression-based methods are a series of models that predict the bounding boxes directly using regression. Liao (\hyperref[Liao2017]{Liao et al.,2017}) design TextBoxes that consist of an end-to-end trainable neural network model for scene text detection. Secondly, propose a word spotting/end-to-end recognition framework that effectively combines detection and recognition. Thirdly, designing a model
achieves highly competitive results while keeping its computational efficiency. Liao (\hyperref[Liao2018]{Liao et al.,2018}) designed TextBoxes++ which directly predicts arbitrary-oriented word bounding boxes by quadrilateral regression to achieve state-of-the-art performance with high efficiency for both horizontal and oriented text. Liao (\hyperref[Liao2018+]{Liao et al.,2018}) uses a method called Rotation sensitive Regression Detector (RRD) Which consists of two network branches of different designs. Concretely, the regression branch extracts rotation-sensitive features by actively rotating the
convolutional filters, while the classification branch extracts rotation-invariant features by pooling the rotation-sensitive features these methods achieve state-of-the-art performance on multi-oriented and long text instances. He (\hyperref[He2017]{He et al.,2017}) uses a text attention mechanism to make an accurate text detector that predicts word-level bounding boxes in one shot. Zhou (\hyperref[Zhou2017]{Zhou et al.,2017}) developed a pipeline that uses pixel-level regression that directly predicts words or text lines of various orientations and quadrilateral forms within complete images, removing the need for redundant intermediate processes (such as candidate aggregation and word segmentation), using a single neural network. Xie (\hyperref[Xie2019]{Xie et al.,2019}) present a novel dimension-decomposition region proposal network (DeRPN) that utilizes an anchor string mechanism to independently match object
widths and heights, which is conducive to treating variant object shapes by preventing the small items from being overshadowed by the larger ones. Despite the efficiency of the regression-based method, It has difficulty handling complex, nonlinear data, is affected by outliers, and relies on the assumptions of independence, homoscedasticity, and normality, which may not be applicable in numerous real-world situations.\\
Sequence-based methods typically detect text by dealing with them as a sequence of characters or word parts. Xie (\hyperref[Kantipudi2021]{Kantipudi et al.,2021}) This research paper proposed an approach for scene text recognition that integrates bidirectional LSTM and deep convolution neural networks. In the proposed method, first, the contour of the image is identified, and then, it is fed into the CNN. CNN is used to generate the ordered sequence of the features from the contoured image. The sequence of features is now coded using the Bi-LSTM. Bi-LSTM is a handy tool for extracting the features from the sequence of words. Thus, this paper combines the two powerful mechanisms for extracting the features from the image and contour-based input image making the recognition process faster. Despite the Sequence-based method's effectiveness, it requires a lot of complexity and time to use LSTMs, GRUs, or Transformers also depends on the order of entered data which makes it very hard to use.
\\
Segmentation-based methods typically merge pixel-wise prediction with post-processing techniques to obtain the bounding boxes. Zhang (\hyperref[Zhang2016]{Zhang et al.,2016}) presented a novel framework for multioriented scene text detection. The main idea that integrates semantic labeling by FCN and MSER provides a natural solution for handling horizontal and multi-oriented text. Wang (\hyperref[Wang2019]{Wang et al.,2019}) proposes a new method called Progressive Scale Expansion Network (PSENet), which can precisely detect text instances with arbitrary shapes. By generating different scales of kernels for each text instance and gradually expanding the minimal scale kernel to the text instance with the complete shape. Xue (\hyperref[Xue2018]{Xue et al.,2018}) introduces a scene text detection method that utilizes semantics-aware text borders and a bootstrapping approach for augmenting text segments. By incorporating semantics-aware text boundaries, the technique significantly enhances the accuracy of localizing text borders with varying meanings. Additionally, the augmented text line segments contribute to a more consistent output of predicted feature maps, resulting in more comprehensive and intact scene text detections. Tian (\hyperref[Tian2019]{Tian et al.,2019}) designed a method that maps pixels onto an embedding space where pixels belonging to the same text are encouraged to appear closer to each other and vice versa to make pixels appear in clusters. Our approach emphasizes enhancing segmentation outcomes by integrating the binarization process during the training phase, all while maintaining inference speed.\\
Most fast-scene text detection methods can not deal with text instances of irregular shapes,
such as curved shapes. Compared to the previous fast scene text detectors, our method not only runs faster but also can detect text instances of arbitrary shapes. Our proposed approach performs more accurately and more efficiently owing to the simple and efficient differentiable binarization algorithm.

\subsection{Optical Character Recognition}
Arabic handwritten text recognition is a particular challenge due to the cursive nature of the script, contextual letter forms, and the small amount of labeled datasets. In the last couple of decades, research has evolved from handcrafted feature-based systems to sophisticated deep learning architectures, driven by the availability of increased computational power and datasets. Below, we survey key methodologies, datasets, and architectural innovations in Arabic Handwritten Text Recognition.

Early AHTR systems relied on manual feature extraction and statistical models.(\hyperref[AlHajj2009]{Al-Hajj et al., 2009})
 combined structural features, such as loops and diacritics, with HMMs to achieve 92.1\% character accuracy on a proprietary dataset. Similarly, (\hyperref[Elzobi2013]{Elzobi et al. 2013}) used Gabor filters to capture texture-based features on the IFN/ENIT dataset (\hyperref[Pechwitz2002]{Pechwitz et al., 2002}), which is a benchmark of 26,459 handwritten Tunisian town names, training SVMs thereon. While these achieved a certain degree of success, their explicit-segmentation reliance and sensitivity to intra-writer variability limited the generalization capabilities.

Deep learning changed the paradigm towards E2E systems. \hyperref[Graves2009]{Graves et al. (2009)} pioneered the CNN-BLSTM-CTC architecture, where a CNN for spatial feature extraction is combined with a BLSTM network for sequence modeling and CTC for label alignment. This framework is particularly effective for cursive scripts like Arabic. \hyperref[Fasha2020]{Fasha et al. (2020)} adopted this architecture for the KHATT dataset \hyperref[Mahmoud2014]{(Mahmoud et al., 2014)}, reporting 80.02\% CRR for unconstrained paragraphs. The work stressed that BLSTMs are essential for modeling bidirectional contextual dependencies, especially because Arabic is written from right to left.

In the area of Arabic handwritten text recognition, several milestones have been achieved. \hyperref[Fasha2020]{Fasha et al. (2020)} presented the Arabic Multi-Fonts Dataset (AMFDS), a synthetically generated corpus that contains 2 million word images in 18 different fonts. The paper employed a CNN-BLSTM-CTC model, trained on this dataset and yielding an 85.15\% CRR on unseen data, proof that synthetic data can help in solving real-life problems suffering from labeled sample scarcity. However, their model gave way to difficulty for some ligatures and intersecting strokes that are very normal in handwritten Arabic texts.

In another similar study, \hyperref[Waly2024]{Waly et al. (2024)} presented an end-to-end OCR system for Arabic handwritten documents. Authors proposed a methodology that included Differentiable Binarization and Adaptive Scale Fusion for improving the accuracy of text segmentation. The AMFDS achieved a CRR of 99.20\% and a WRR of 93.75\% on single-word samples containing 7 to 10 characters, as well as a CRR of 83.76\% for sentences. These results underpin the robustness of the AMFDS for training models on high-accuracy recognition tasks.

Hybrid models and data augmentation techniques are some other methods that have been tried to increase the recognition rate.\hyperref[Gresha2023]{Gresha et al. (2023)} propose a hybrid deep learning approach for handwritten Arabic text recognition, using a combination of CNNs with Bidirectional Recurrent Neural Networks, specifically Bidirectional Long Short-Term Memory (Bi-LSTM) and Bidirectional Gated Recurrent Units (Bi-GRU). This model combines the power of capturing spatial features with that of modeling temporal dynamics and contextual relationships in handwritten Arabic text.

Following this, experiments on the Arabic Handwritten Character Dataset(AHCD)  and the Hijjaa benchmark datasets were conducted to demonstrate the effectiveness of the proposed hybrid models. In fact, for the very first time, the CNN-Bi-GRU framework outperformed state-of-the-art accuracy rates at 97.05\% and 91.78\% on the AHCD and Hijjaa datasets, respectively, compared with already proposed deep learning-based methods. These results emphasize the notable performance increase due to the incorporation of specialized temporal modeling and contextual representation capabilities within the handwriting recognition pipeline with no explicit segmentation.

Besides, it has also generated some promise regarding data augmentation with the integration of GANs. A work by \hyperref[Gresha2023]{Gresha et al. (2023)} integrated GANs within a CNN-BLSTM architecture for Arabic handwritten recognition and reported a recognition rate of 95.23\% on the \hyperref[Pechwitz2002]{Pechwitz2002}. This approach showed that GANs could generate realistic synthetic samples effectively to augment the training dataset and thus improve model performance.
The attention mechanism has contributed much to the evolution of Arabic Handwritten Text Recognition. It helps the network give more attention to the pertinent region within the text. Introduced for neural machine translation by \hyperref[Bahdanau2015]{Bahdanau et al. (2015)}, it finds applications in many domains such as OCR.

\hyperref[Gader2022]{Gader and Echi (2022)} proposed an attention-based CNN-ConvLSTM model for extracting handwritten Arabic words. The developed approach succeeded in extracting words at a rate of 91.7\% in the \hyperref[Khatt2016]{KHATT dataset}, demonstrating in this way how attention mechanisms improved recognition accuracy.

The Transformer model, first proposed by \hyperref[Vaswani2017]{Vaswani et al. (2017)}, has drastically improved several aspects of NLP through its self-attention mechanism, which efficiently captured long-range dependencies. For AHTR, the transformer models were used to handle the sequential nature of the handwriting without recourse to any form of recurrent architecture.

\hyperref[Momeni2023]{Momeni and BabaAli (2023)} proposed a transformer-based model for offline Arabic handwritten text recognition. The methodology has utilized the self-attention mechanism of transformers, which is useful in capturing the long-range dependencies of the text as one of the challenges posed by the cursive nature of Arabic script.
It is based on a pre-trained transformer for both image understanding and language modeling. Processing text line by line, the overall context of the handwritten text is taken into consideration, which is quite valuable in Arabic; the shape of any character can change according to its position inside a word. Therefore, line-level processing allows the model to deal with the peculiarities in Arabic handwriting and increase recognition accuracy.
It was evaluated on the Arabic KHATT dataset-a well-known benchmark for Arabic handwritten text recognition. The obtained results evidence that the proposed transformer-based approach clearly outperforms the performance of the traditional methods and reaches an CER of 18.5\%, thus proving the effectiveness of transformer architecture in the modeling of complex patterns of Arabic handwriting.

Scarcity of labeled datasets in AHTR has thus induced the employment of pre-trained models and transfer learning techniques. AraBERT is a pre-trained transformer model for Arabic text, developed by \hyperref[Antoun2020]{Antoun et al. (2020)}. It is trained on a large corpus of Arabic text and has been applied to several downstream tasks, including AHTR, in order to improve recognition accuracy.

Qalam realizes a key landmark in development upon the success in AHTR advancements. Qalam is a large multimodal model for Arabic Optical Character Recognition (OCR) and Handwriting Recognition (HWR). Qalam's architecture is an ensemble of the SwinV2 encoder \hyperref[Liu2022]{Liu et al., 2022} and the RoBERTa decoder \hyperref[Liu2019]{Liu et al., 2019} model architecture to fully capture the different patterns and peculiarities present within Arabic.

Qalam, trained on the diverse dataset of more than 4.5 million images of Arabic manuscripts \hyperref[Manuscripts2023]{Al-Mutawa et al., 2023} and a synthetic dataset of 60,000 image-text pairs, has exceptional performance. It achieves a WER of 0.80\% in HWR tasks and 1.18\% in OCR tasks, outperforming existing methods by a wide margin \hyperref[Qalam2024]{Qalam, 2024}. More importantly, Qalam handles Arabic diacritical marking with unparalleled capability and processes high-resolution inputs nicely, two common weaknesses in current OCR systems.

These developments further illustrate how Qalam can be at the vanguard in Arabic script recognition, providing that leap in terms of accuracy and efficiency for applications both historical and modern.

\section{Experiments}
Given the limited resources for training large complex models, we tried out different architectures to reach the best solution possible. We tried out 4 different models with the same data and training technique.
\subsection{CNN-BLSTM-CTC}
Initially, we tried out this architecture as CNN acts as a Feature Extractor that would be further processed via the Recurrent Bidirectional Network of LSTM and then converted to characters by Connectionist Temporal Classification. Our objective was to implement a base solution for the recognition stage before we could optimize or fix issues related to the overall solution with expectations to approach the accuracy of existing solutions with that architecture done on English Handwritten.
\begin{figure*}[ht]
    \centering
    \includegraphics[width=\textwidth]{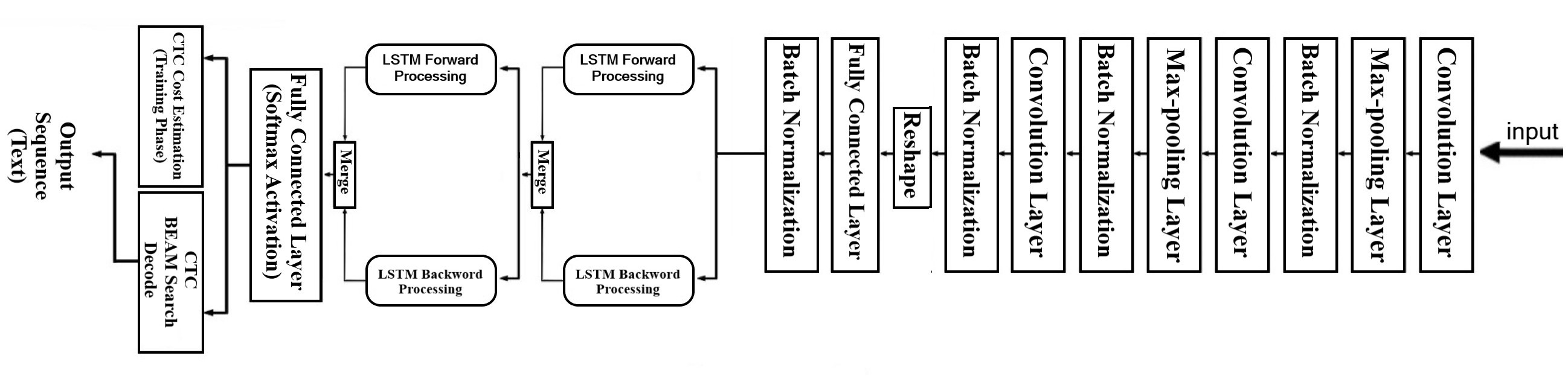}
    \caption{An illustrative visualization of The OCR architecture that begins with the input image being fed into the Convolutional Neural Network (CNN) module for feature extraction, capturing spatial patterns and visual features. The extracted feature maps are then processed into a sequence of feature vectors which serve as input to the Bidirectional Long Short-Term Memory (BLSTM) network. Finally, the output is passed through the Connectionist Temporal Classification (CTC) layer, which decodes the sequence into readable text.}
    \label{fig: CNN-BLSTM-CTC OCR architecture}
\end{figure*}
\begin{figure}
    \centering
    \includegraphics[width=0.999\linewidth]{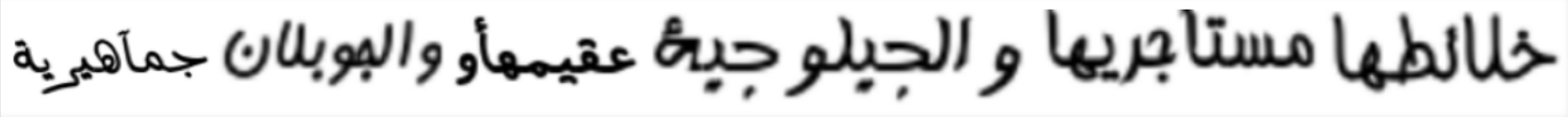}
    \caption{Generated line from the raw words data in (AMFDS)}
    \label{fig: Sample generated line from AMFDS}    
\vspace{-17pt}
    
\end{figure}

This model is designed to handle the Segmented Arabic Line of text. The detailed architecture is described in Table \ref{tab:model-CNN-BLSTM--CTC-architecture} and visualized in Figure \ref{fig: CNN-BLSTM-CTC OCR architecture}. The model was trained on a data generator that simulates a randomized, synthesized dataset with augmentation applied at random and a specified ratio of non-augmented data to augmented data.\\
The creation of sentence lines was done using the Arabic Multi-Fonts Dataset (AMFDS) (\hyperref[Fasha2020]{Fasha et al.,2020}) as base row data of words that later will be concatenated randomly to form a sentence with a controlled maximum number of generated words per line as in the Figure \ref{fig: Sample generated line from AMFDS}. As it appears in the figure of the generated line, all the words are scaled to a fixed height by locking the aspect ratio and then concatenated to form the desired line before rescaling the whole image of the text line to fit the width of 1024 given that we set the height of the input to 64. If the width is smaller than the input width and the height equals the input height we pad the left side of the image until reaching the input width. If the width exceeds the input's, we scale the image down until it reaches the desired width by centering the line vertically and putting the remaining height if needed.
\vspace{-10pt}

\begin{flushright}
\begin{table}[ht]
\vspace{-10pt}
\caption{Detailed architecture of the OCR Model}
\hspace*{-0.3cm}
\label{tab:model-CNN-BLSTM--CTC-architecture}
\centering
\begin{tabular}{|c|c|c|}
\hline
\textbf{Layer} & \textbf{Config} & \textbf{Notes} \\
\hline
Conv layer& $3 \times 3$, 32 & Activation: ReLU\\
\hline
MaxPooling & $2 \times 2$ & Pooling window \\
\hline
BatchNorm & - & - \\
\hline
\hline
Conv layer& $3 \times 3$, 64 & Activation: ReLU\\
\hline
MaxPooling& $2 \times 2$ & Pooling window \\
\hline
BatchNorm & - & - \\
\hline
\hline
Conv layer& $3 \times 3$, 128 & Activation: ReLU\\
\hline
BatchNorm & - & - \\
\hline
\hline
Dense & 64 units & Activation: ReLU \\
\hline
BatchNorm & - & - \\
\hline
\hline
Bi-LSTM & 128 units & Return sequences \\
\hline
Bi-LSTM & 256 units & Return sequences \\
\hline
\hline
Dense & Softmax & Classes: \# vocabulary + [blank] \\
\hline
\hline
CTC & - & Connectionist Temporal Classification \\
\hline
\end{tabular}
\end{table}

\vspace{-17pt}
\end{flushright}

In our experiments with this model, with a learning rate of \(5 \times 10^{-5}\) the model fails to converge on long sequences right from the beginning. To solve this issue, we trained the model over 2-3 word sentences then applied transfer learning techniques with freezing the Convolutional Neural Network "Feature Extraction" and kept on training the rest of the architecture. This approach was found successful with a reasonable decline in accuracy with each increase in the word count per sentence. However a major problem has come up with this model architecture, and that is the accuracy declines rapidly on long sequences as shown in Table \ref{tab:OCR model-CNN-BLSTM-CTC-Performance}. In the results table, WRR values are very low compared to the CRR and that's due to the sampling dataset having 18 fonts that mimic the handwritten with a range of 7-10 characters per word. The Table shows the results obtained on sequence length from 1 to 6 words as a longer sequence would result in a very bad accuracy with half of the sentence not recognized. Yet the accuracy is not very sufficient in terms of real-world application as it will require a significant amount of manual modification on the OCR result. 
\begin{table}[h!]
\vspace{-7pt}
\centering
\caption{OCR Model CNN-BLSTM--CTC Performance Results}
\label{tab:OCR model-CNN-BLSTM-CTC-Performance}
\vspace{-7pt}
\begin{tblr}{
  colspec = {|c|c|c|c|},
  hlines,
  vlines,
  colsep = 2pt,
  rowsep = 0.8pt
}
\# & Num of Words & CRR (\%) & WRR (\%) \\ 
1  & 1 & 99.20 & 93.75 \\ 
2  & 2 & 93.67 & 62.18 \\ 
3  & 3 & 93.70 & 60.20 \\ 
4  & 4 & 84.15 & 31.42 \\ 
5  & 5 & 81.29 & 30.87 \\ 
6  & 6 & 66.01 & 18.78 \\ 
\end{tblr}
\vspace{-7pt}
\end{table}
\\    
With these problems, we decided to consider another alternative or to modify the recurrent network in that approach to overcome the mentioned issues.

After defining all the defects of the previous approach we tried out in parallel 3 different approaches. One was to try debugging the issues found in the previous approach by replacing BLSTM with  Mini-LSTM to accelerate the training rapidly and add self-attention layer. The 2 other approaches were trying out most 2 other architectures that seemed very promising on the OCR task with more advanced decoding and mapping of features. Those approaches are Encoder-Decoder Vision Transformer \& CNN-Transformer Architecture.
\subsection{CNN-Mini-LSTM-SelfAttention-CTC}
We applied two changes to CNN-BLSTM--CTC model tring to improve its results.\\
1- use miniLSTM instead of using LSTM which has problems in longer sequences, as it has a quadratic computational complexity with respect to sequence length, this means that it takes a lot of time and resources to capture complex dependencies.
miniLSTM has a simplified architecture by Removing Hidden State Dependencies, Dropping tanh and sigmoid Constraints, and Ensuring Output Independence.\\
2- use self-attention Layer which is a powerful part of deep learning that lets a model figure out how vital each sequence part is. This lets the model better understand dependencies and relationships in the data.\\
so the new model architecture the same as the one in Table \ref{tab:model-CNN-BLSTM--CTC-architecture}  but with two major different every BLSTM layer changes with miniLstm and a self attention layer added before last desnse layer.\\
However the problems in the CNN-BLSTM--CTC model still appear in this model architecture but the results slightly improved as the accuracy declines slowly on long sequences compared to the CNN-BLSTM--CTC model but we face a new problem in the CTC decoder as it decode the word with the additional Arabic character which make the accuracy still not very sufficient in terms of real-world application as it will require a significant amount of manual modification on the OCR result.
\subsection{Vision Encoder-Decoder Transformers}
\begin{figure}
    \centering
    \includegraphics[width=0.999\linewidth]{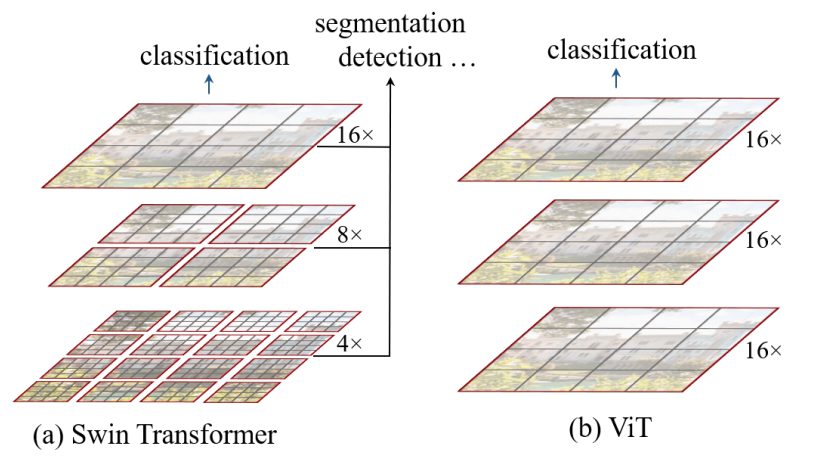}
    \caption{The Swin Transformer builds hierarchical feature maps with linear complexity by computing self-attention within local windows, unlike regular vision Transformers with single low-resolution maps and quadratic complexity.}
    \label{fig: Swin-Vit-comparison}    
\vspace{-17pt}
\end{figure}

We considered Vision Encoder-Decoder Transformer architecture. The selection of the encoder and decoder was done given the performance results for each encoder and decoder type on the Arabic OCR task (\hyperref[Bhatia2024]{Bhatia at el.,2024}). The top performing Encoder is Swin V2 according to the study and AraBERT got the finest results although they chose to make RoBERTa their final decoder due to a long spacing recognition. The Swin Transformer is a versatile computer vision backbone that has demonstrated exceptional performance across a range of tasks, including region-level Object Detection, pixel-level Semantic Segmentation, and image-level Classification. Its core innovation lies in incorporating essential visual priors such as hierarchical structure, locality, and translation invariance into the standard Transformer Encoder as shown in Figure \ref{fig: Swin-Vit-comparison}. Performing the previous mechanism in addition to The shifted window mechanism in Swin V2 enhances feature interaction across windows by introducing overlapping attention regions, improving context modeling while maintaining efficiency. The overall approach combines the powerful modeling capabilities of the Transformer with visual priors that enhance its adaptability and effectiveness in diverse visual tasks.
\\
Following the training methodology applied to the state-of-the-art method of initializing the training weight with an Encoder trained on image classification task and a Decoder trained on text generation (\hyperref[Li2023]{Li et al.,2023}).
The Encoder initial weights were the best weights on the ImageNet image classification dataset (\hyperref[Russakovsky2015]{Russakovsky et al.,2015}). On the other hand, the Decoder's initial weights were the official weights of AraBERT which were trained on various Arabic content (\hyperref[Antoun2020]{Antoun et al.,2020}).

We tested the architectures initially on the generated dataset created from AMFDS and then on the Naqsh Arabi OCR Dataset (our dataset) as described in \ref{methodology Section} and the publicly available Online Khatt dataset. \\
The test was done on the dataset generated by AMFDS. the test contained 100 batches with a batch size of 32 line images. Each line contained 20 randomly generated words. All the tests appear in Table\ref{tab:OCR model-bert-swinv2-Performance}. 
\begin{table}[h!]
\vspace{-7pt}
\centering
\caption{OCR Model BERT-SwinV2 Performance Results}
\label{tab:OCR model-bert-swinv2-Performance}
\vspace{-7pt}
\begin{tblr}{
  colspec = {|c|c|c|c|},
  hlines,
  vlines,
  colsep = 2pt,
  rowsep = 0.8pt
}
\# & Dataset & CRR (\%) & WRR (\%) \\ 
1  & Generated Dataset from AMFDS [\ref{Waly2024}]& 94.34 & 70.03 \\ 
2  & Naqsh Arabi OCR Dataset(Ours) & 96.45 & 92.08 \\ 
3  & Online Khatt [\ref{Mahmoud2018}]& 74.91 & 50.34 \\
\end{tblr}
\vspace{-7pt}
\end{table}\\
From the results previously mentioned we noticed that the compression of the images to 256x256 caused obvious information loss which led to non-optimal performance compared to a recent study of Qalam (\hyperref[Bhatia2024]{Bhatia et al.,2024}) with experiments on the same model on Khatt and online Khatt datasets with input size of 1024x1024 which performed better.

\section{Methodology} \label{methodology Section}
\subsection{Detection Module}
Our work Method initially Follows the recently proposed DBNet++ (\hyperref[Liao2023]{Liao et al.,2023}) which performed well on several printed text datasets. As our detection objective "Handwritten Arabic Text" is similar to the printed text, we fine-tuned Universal best-weights trained on \hyperref[ICDAR2015]{ICDAR 2015}, \hyperref[TotalText2020]{Total-Text},  \hyperref[TD500]{MSRA-TD500}, and \hyperref[Zhang2021]{Chinese Baidu} using Handwritten text images to add value to the universal weights.\\
1- Model background (DBNet++) (\hyperref[Liao2023]{Liao et al.,2023}):
The model starts by feeding the image into the ResNet50 Backbone for Feature Extraction then scales them up until to reach the same scale for all to pass the features to The Adaptive Scale Fusion (ASF) module. The ASF generates contextual features to predict probability map and threshold map as in Figure 
After that comes the calculation of Approximate Binary maps using a Probability map and a Feature map. During the training period, the supervision is applied on the Probability map, the Threshold
map, and the Approximate Binary map, where the Probability map and the Approximate Binary map share the same Supervision. 
On Testing, Bounding Boxes are derived from either the approximate Binary map or the Probability map via a box formation process.\\ 
2- Segmentation Dataset:
We used Arabic Documents OCR Dataset (\hyperref[Arabic2023]{Arabic2023}) for our Line Segmentation Task.
The dataset contains 10K printed and handwritten text images split into 12 classes. We only used two classes of the dataset "Handwritten text and official documents" Which suit our problem of 1.6K images.

\subsection{OCR Module}
We have created an Optical Character Recognition (OCR) module which can identify various types of texts from images. This includes:
\begin{itemize}
    \item \textbf{Handwritten Text:} Writings and notes of various individuals, each with his/her own writing style.
    \item \textbf{Printed Text:} Standard text found in books, documents, and printed materials.
    \item \textbf{Arabic Numerals:} Numbers written in Arabic script, both printed and handwritten.

\end{itemize}
In order to offer accurate and reliable text recognition across these formats, we have combined two robust technologies:
\begin{enumerate}

\item \textbf{The Convolutional Neural Network (CNN):} feature extractor is the core element of the module used to process images for recognizing the key visual features,This ability enables the system to "understand" the text in the images.
\item \textbf{Transformer-Based Encoder-Decoder Architecture:} It converts the characteristics that have been learned by the CNN into the correct sequence of characters. It enables the system to "interpret" and reconstruct text from visual input.
\end{enumerate}
With these technologies combined, our OCR module is capable of reading and interpreting text in different forms at high accuracy levels, thus digitizing images to editable and searchable information.

\subsection{Datasets}
For training and testing our OCR module, we have used a number of datasets that include a variety of texts. These indeed include images of handwritten texts, printed texts, and also Arabic numerals. Each dataset is discussed separately below.

\textbf{{1) Online KHATT Dataset}} \\
In our work, the Online KHATT Dataset is used for handwritten text recognition. It is one of the large datasets of Arabic handwritten text samples and finds wide application in research related to Arabic handwriting recognition.

\begin{itemize}
\item \textbf{Training Samples}: 6,974 images
\item \textbf{Validation Samples}: 767 images
\item \textbf{Test Samples}: 766 images
\end{itemize}

The Online KHATT dataset is composed of writings from many individuals, which means variation in handwriting style, slant, and pressure. This is important for developing a robust handwriting recognition system that can be generalized to variations that may come in real-world scenarios.

\textbf{{2) Naqsh Arabi OCR Dataset}}\\
In order to optimize accuracy and to make the decoder more familiar with the Arabic language patterns, we created a dataset using the Abstract Window Toolkit graphics Java package, and the content was extracted from the Arabic content on Wikipedia
The created dataset is a line-synthesized Arabic OCR dataset that contains 21 different calligraphy forms of different fonts with some fonts existing with various weights.\\
As shown in Table \ref{tab:font_table}, the dataset includes sample images of line text and their corresponding font names.

The text data used to create the dataset was extracted from several hundreds of Wikipedia websites each categorized in one of the sciences shown in Table \ref{tab:Breakdown of Wikipedia pages extracted by category}. The details of the extracted pages URLs can be found in the Appendix \ref{Naqsh Dataset Appendix}.
\begin{table}[ht]
\centering
\renewcommand{\arraystretch}{1.8} 
\setlength{\tabcolsep}{3pt} 

\caption{Table showing images of text lines and their corresponding fonts.}
\label{tab:font_table}
\begin{tabular}{|c|c|}
\hline
\textbf{Sample Image} & \textbf{Font Name} \\ \hline
\includegraphics[width=4cm]{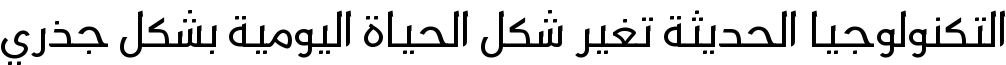} & Abd ElRady Thin \\ \hline
\includegraphics[width=4cm]{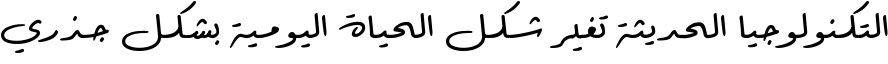} & (A) Arslan Wessam A \\ \hline
\includegraphics[width=4cm]{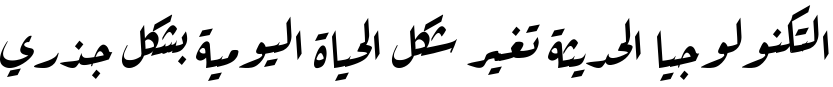} & Aref Ruqaa Regular \\ \hline
\includegraphics[width=4cm]{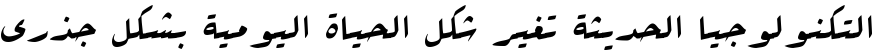} & BArabicStyle \\ \hline
\includegraphics[width=4cm]{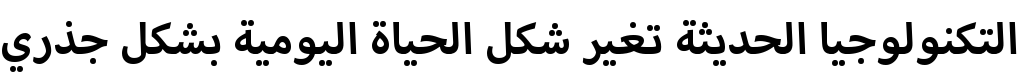} & Bahij Myriad Arabic Bold \\ \hline
\includegraphics[width=4cm]{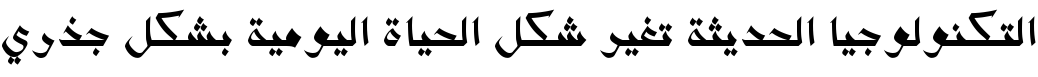} & HSN Omar \\ \hline
\includegraphics[width=4cm]{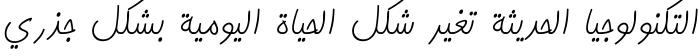} & K Kamran \\ \hline
\includegraphics[width=4cm]{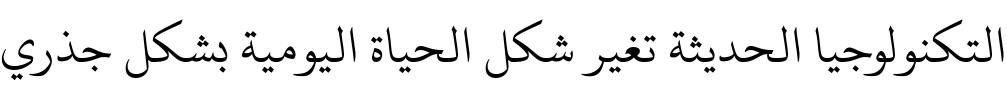} & KFGQPC Uthman Taha Naskh \\ \hline
\includegraphics[width=4cm]{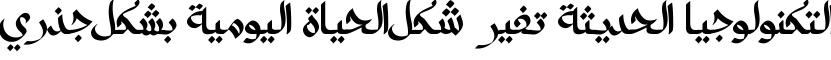} & Khediawy \\ \hline
\includegraphics[width=4cm]{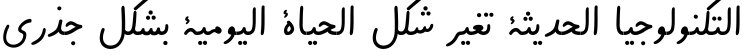} & Mj\_Faraz \\ \hline
\includegraphics[width=4cm]{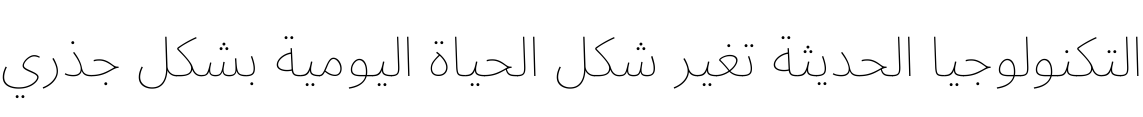} & Noto Sans Arabic Thin \\ \hline
\includegraphics[width=4cm]{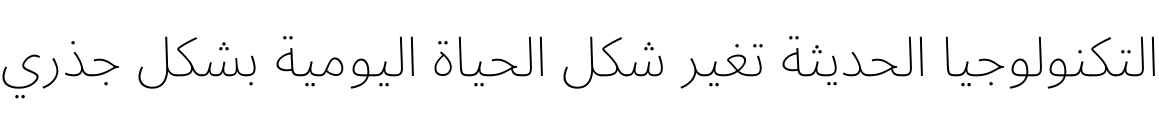} & Noto Sans Arabic ExtraLight \\ \hline
\includegraphics[width=4cm]{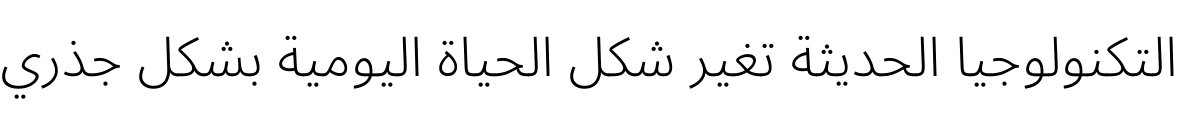} & Noto Sans Arabic Light \\ \hline
\includegraphics[width=4cm]{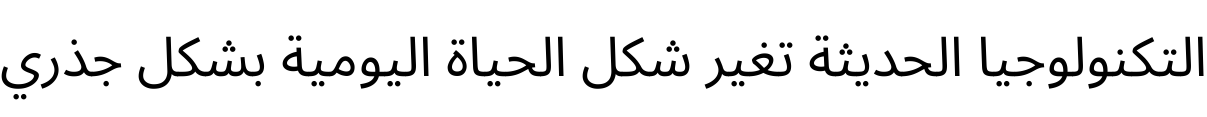} & Noto Sans Arabic Regular \\ \hline
\includegraphics[width=4cm]{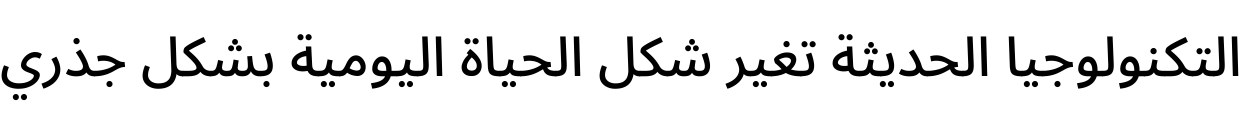} & Noto Sans Arabic Medium \\ \hline
\includegraphics[width=4cm]{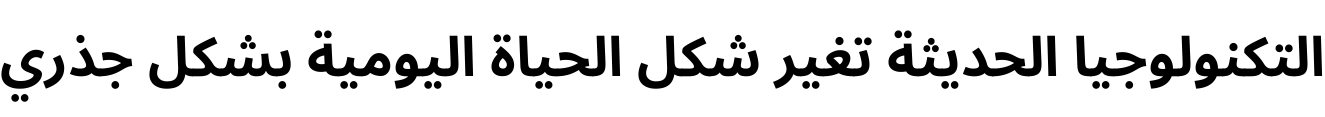} & Noto Sans Arabic Bold \\ \hline
\includegraphics[width=4cm]{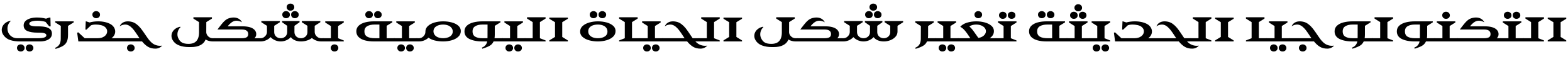} & RTL-Elegance Ar Pro Black \\ \hline
\includegraphics[width=4cm]{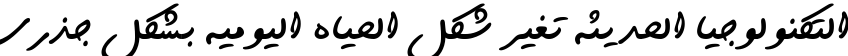} & pHalls Khodkar \\ \hline
\includegraphics[width=4cm]{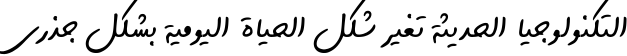} & Shekari Font \\ \hline
\includegraphics[width=4cm]{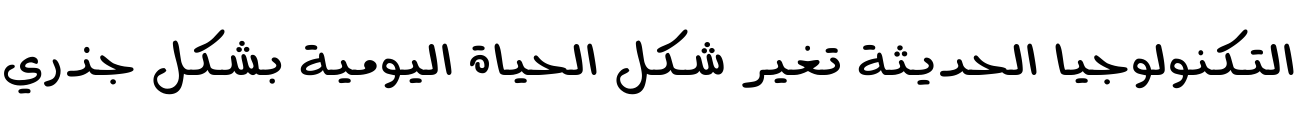} & W\_qalam \\ \hline
\includegraphics[width=4cm]{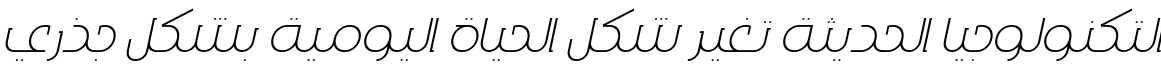} & VEXA thin italic R \\ \hline
\end{tabular}
\vspace{-7pt}
\end{table}

After the Extraction of a text content filtration was required to filter out unwanted symbols and to ensure Arabic existed only in the dataset so we did the following:
\begin{itemize}
\item First Remove the heading and footer of each page to retrieve only the content of the page.
\item Then converted every English digit to Arabic digits.
\item Removing all English words that may be present in the extracted content.
\item Removing diacritics "Tashkeel" as our problem lies with recognizing the characters alone, not with diacritics.
\item Filtering out all symbols that can't be found in a regular Arabic written word. (As we're targeting a real-world application we assumed that the presence of symbol recognition will be very beneficial especially if used to recognize text in a financial context)
\end{itemize}

\begin{figure}[!hbtp]
    \centering
    \includegraphics[width=0.999\linewidth]{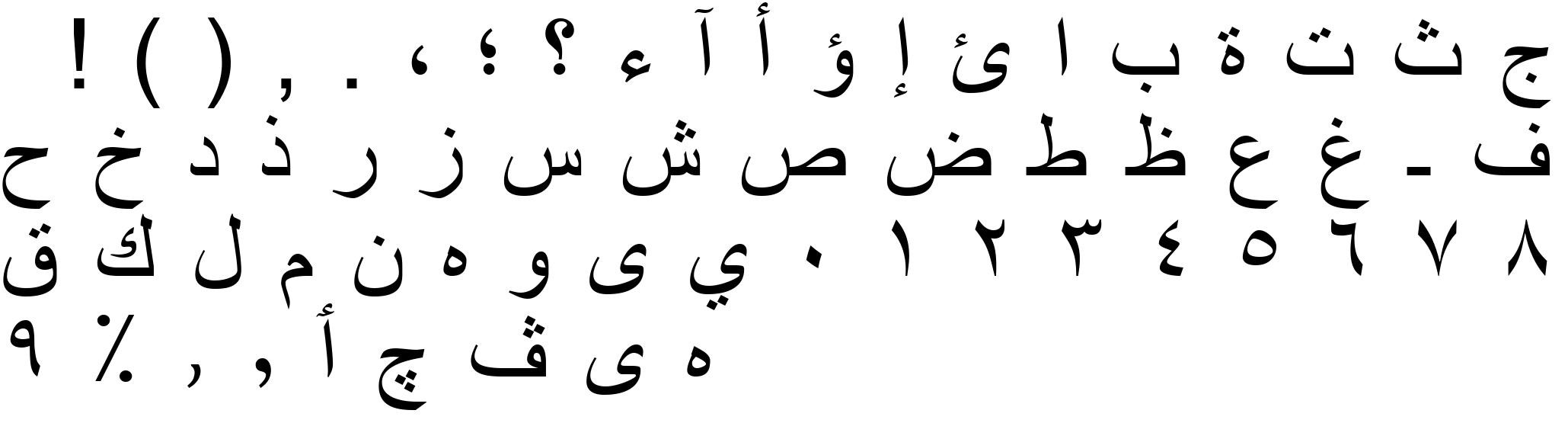}
    \caption{The Allowed character set of the Dataset extracted Text Content that consists of 64 characters, digits, and symbols}
    \label{fig: allowed charset}
\vspace{-7pt}
    
\end{figure}

\begin{table}[!hbtp]
\centering
\caption{Breakdown of Wikipedia pages extracted by category}
\label{tab:Breakdown of Wikipedia pages extracted by category}
\begin{tabular}{|>{\centering\arraybackslash}c|>{\centering\arraybackslash}c|}
\hline
\textbf{Category} & \textbf{Number of Pages} \\ \hline
Geography & 110 \\ \hline
History & 115 \\ \hline
Information Technology & 111 \\ \hline
Poetry & 123 \\ \hline
Mathematics & 112 \\ \hline
Languages & 108 \\ \hline
Politics & 116 \\ \hline
Accounting & 111 \\ \hline
Commerce & 120 \\ \hline
Industry & 110 \\ \hline
Philosophy & 116 \\ \hline
War & 122 \\ \hline
\end{tabular}
\vspace{-10pt}
\end{table}
This dataset comprises:
\begin{itemize}
\item \textbf{Training Samples}: 84,000 images
\item \textbf{Validation Samples}: 24,000 images
\item \textbf{Test Samples}: 24,000 images
\end{itemize}
Each is an image of a line of text varying in length from 1 to 20 Words.

\textbf{{3) Arabic Numerals Dataset}} \\
The Arabic Numerals Dataset consists of 7,000 images of handwritten digits. The original dataset has huge variations in handwriting styles, stroke thickness, and alignment. Most of the images also suffer from noise, uneven backgrounds, and inconsistent lighting, making the recognition difficult.

Preprocessing to enhance the quality of data was done as follows:
\begin{itemize}
\item Noise Reduction: These are unwanted marks that were removed from the images.
\item Standardized backgrounds: The uniformity of all the image's backgrounds.
\item Enhanced contrast between the digits was enhanced to the best possible value.
\item Size Normalization: Normalisation of the dimensions of all digits.
\end{itemize}

These steps ensured homogeneity and improved the readability of the dataset, making it more suitable for OCR model training.

To simulate real-world applications where multi-digit numbers need to be recognized, individual digit images were combined into random sequences of (1-8 digits). Additionally, visually similar digits were clustered to ensure consistency across the sequences.

\begin{figure}[ht]
\centering
\includegraphics[width=0.5\textwidth]{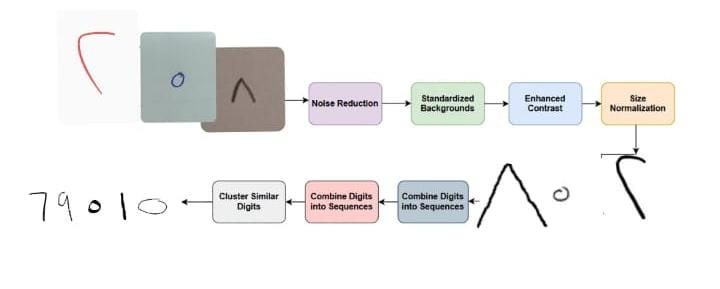} 
\caption{Data Pipeline for Arabic Numerals Dataset Preprocessing. The diagram illustrates the steps involved in combining digit images into sequences and clustering visually similar digits.}
\label{fig:data_pipeline}
\end{figure}

These steps are illustrated in Figure~\ref{fig:data_pipeline}, where the process of combining digits and clustering is shown in detail.

The dataset is split as follows:
\begin{itemize}
    \item \textbf{Training Samples}: 4,900 images
    \item \textbf{Validation Samples}: 1,050 images 
    \item \textbf{Test Samples}: 1,050 images 
\end{itemize}

\subsection{Model Architecture}
\label{sec:architecture}
The architecture of our OCR system is based on a transformer-based model that has been designed for Arabic handwriting recognition. It includes three major components: a Convolutional Neural Network (CNN) for feature extraction, a Transformer encoder to learn the spatial dependencies, and a Transformer decoder for sequence generation. This allows the integration of both spatial features in images and sequential dependencies inherent in Arabic script.

\begin{figure}[ht]
    \centering
    \includegraphics[width=0.55\textwidth]{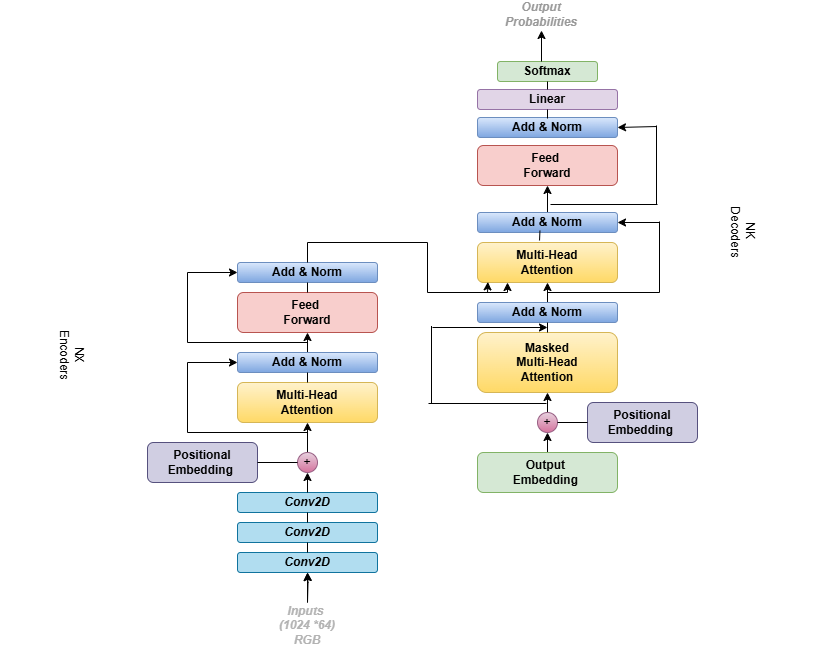}
    \caption{Architecture of the proposed CNN-Transformer model for Arabic OCR. The system processes input images through convolutional layers, encodes spatial features with transformer blocks, and decodes character sequences autoregressively.}
    \label{fig:arch}
\end{figure}

Our model extends the work done by \hyperref[alwajih2022]{Alwajih et al., 2022}, which proposed a transformer-based architecture for Arabic OCR. We expand their architecture with some important changes that enable our model to deal with the challenges of Arabic handwriting in terms of sequence length, positional encoding, and attention mechanisms.
Figure~\ref{fig:arch} shows the overall architecture.

\paragraph{Component Specifications}
\label{subsec:components}

For evaluating the performance of our model, we present its comparison against the baseline and that by \hyperref[alwajih2022]{Alwajih et al., 2022}. A comparison is presented in Table~\ref{tab:arch-comparison} with the following technical enhancements:

\begin{table}[ht]
\centering
\renewcommand{\arraystretch}{1.3}
\caption{Architectural Comparison with Baseline Model}
\label{tab:arch-comparison}
\begin{tabular}{|l|c|c|}
\hline
\textbf{Component} & \textbf{Baseline} & \textbf{Our Model} \\ \hline
CNN Architecture & 2 layers & \textbf{3 layers} \\ \hline
Positional Encoding & Learnable & \textbf{Sinusoidal} \\ \hline
Encoder/Decoder Layers & 6E, 1D & \textbf{6E, 6D} \\ \hline
Attention Heads & 4 per layer & \textbf{8 per layer} \\ \hline
\end{tabular}
\end{table}

\paragraph{(CNN) Feature Extractor}
The Convolutional Neural Network is the first part of the model, and its purpose is to extract the most important visual features from the input image. An RGB image with a size of (1024 x 64) pixels passes through three convolutional layers, each followed by batch normalization and max-pooling operations. These steps contribute to reducing the spatial dimensions, while extracting the features hierarchically from low-level features such as edges and textures to high-level abstract features necessary for the recognition of complex handwriting.

\hyperref[Vaswani2017]{Vaswani et al. (2017)}

\paragraph{Transformer Encoder}
\label{subsec:encoder}

The Transformer encoder takes as input a sequence of features generated by the CNN. It uses multi-head self-attention to model long-range dependencies in the sequence of features. This self-attention allows every step of the model to attend to all parts of the input sequence, thereby enabling it to capture contextual relationships between characters that may be far apart in the input image.
The contribution of our work lies in developing a model incorporating the sinusoidal positional encoding  proposed for the first time in \hyperref[Vaswani2017]{Vaswani et al. (2017)}, part of a larger paper titled \textbf{"Attention is All You Need"}. This kind of encoding will efficiently provide our model with contextual information on character positions, which is important to Arabic handwriting recognition because such letters largely depend on word ordering.
The encoder consists of 6 layers, which allow the model to obtain an increasingly higher level of complexity of the input sequence.

\paragraph{Transformer Decoder}
\label{subsec:encoder}
The Transformer decoder is an autoregressive generation of the output sequence. It generates one character at a time, given previous predictions and the encoder's encoded features for context. In order to prevent "cheating" by looking ahead at the tokens later in the sequence to predict the current one, the model uses masked self-attention.
Our model has a decoder consisting of 6 layers, while \hyperref[alwajih2022]{Alwajih et al., 2022} used 1 decoder layer. This deeper decoder will be able to handle complex character dependencies, especially for ambiguous or overlapping characters.
\paragraph{Positional Encoding}

A main improvement of our model is that sinusoidal positional encoding is utilized. Unlike the learnable position embeddings, the sinusoidal encoding relies on a fixed function. It is proven to be much more effective when dealing with very long sequences. This encoding informs the model the relative positions of characters in one sequence, which is crucial in handwriting recognition.
\begin{equation}
\begin{split}
\text{PE}_{(pos,2i)} &= \sin\left(\frac{pos}{10000^{2i/d_{\text{model}}}}\right), \\
\text{PE}_{(pos,2i+1)} &= \cos\left(\frac{pos}{10000^{2i/d_{\text{model}}}}\right)
\end{split}
\end{equation}

where $d_{\text{model}}=256$ is the embedding dimension and $pos$ denotes the character position. This sinusoidal encoding, as originally proposed in \hyperref[Vaswani2017]{Vaswani et al. (2017)}, was used in place of learnable embeddings that used in \hyperref[alwajih2022]{Alwajih et al., 2022}
\paragraph{Technical Configuration}

Table~\ref{tab:config} lists hyperparameters optimized for Arabic handwriting recognition. Smaller dropout and using AdamW optimization are among the key decisions.

\begin{table}[ht]
\centering
\caption{Model Hyperparameters from Configuration}
\label{tab:config}
\begin{tabular}{|l|c|}
\hline
\textbf{Parameter} & \textbf{Value} \\ \hline
Embedding dimension ($d_{\text{model}}$) & 256 \\ \hline
Encoder layers & 6 \\ \hline
Decoder layers & 6 \\ \hline
Attention heads & 8 \\ \hline
Feedforward dimension & 512 \\ \hline
Dropout rate & 0.1 \\ \hline
Batch size & 16 \\ \hline
Base learning rate & 0.0001 \\ \hline
Training epochs & 55 \\ \hline
\end{tabular}
\end{table}

\paragraph{Performance Improvements}
\label{subsec:performance}

Based on these architectural enhancements, our proposed model presents quite a drastic improvement in handling the challenges of Arabic handwritten text. These enhance efforts significantly optimize the model's performance in terms of accuracy, robustness, and efficiency, especially with diverse handwriting styles and longer sequences of text.

Table~\ref{tab:improvements} summarizes the most important architectural improvements and the respective impact on the model's performance. These will enable the system to recognize Arabic handwriting more precisely, especially under challenging conditions like connected letters, long sequences, and varying styles of handwriting.

\begin{table}[ht]
\centering
\renewcommand{\arraystretch}{1.4} 
\setlength{\tabcolsep}{6pt} 
\caption{Performance Gains from Architectural Enhancements}
\label{tab:improvements}
\small
\begin{tabular}{|p{2cm}|p{5cm}|}
\hline
\textbf{Enhancement} & \textbf{Impact on Arabic OCR} \\ \hline
\textbf{Additional CNN layer with batch normalization \& Max pooling} & Adding an extra convolutional layer with batch normalization and max pooling keeps the fine-grained stroke level of detail very important in the recognition of Arabic ligatures. Max pooling decreases the spatial dimensions, retaining the most important features and reducing computation; hence, the structures of connected letters will be more clear in recognition. \\
\hline
\textbf{8 attention heads per layer} & Enhances the ability to model long-range dependencies in Arabic cursive script, improving overall recognition accuracy. \\
\hline
\textbf{Sinusoidal positional encoding} & Provides improved handling of long sequences (up to 2048 characters), making the OCR system more effective for extended text inputs. \\
\hline
\textbf{Multi-dataset training} & Strengthens model robustness, allowing it to generalize better across different handwriting styles and real-world variations. \\
\hline
\end{tabular}
\end{table}

\paragraph{Technical Details}
\label{sec:technical}
1. \textbf{Real-World Applicability}\\
Our OCR system aims to recognize Arabic printed and handwritten numbers on receipts and apply this technology in real-world applications. In order to make our system more robust and adaptable, we extended our dataset by adding images with different distortions that can be found in scanned receipts.

2. \textbf{Augmentation}
To make the model generalize well for real-world conditions, we trained the model on images that include:
\begin{itemize}
\item Background noise such as lined paper and printed receipt textures.
\item Motion blur and low resolution to simulate scanning imperfections.
\item Rotations and distortions reflecting handwritten variations.
\item Salt-and-pepper noise and Gaussian noise to account for scanner artifacts.
\end{itemize}

\begin{figure}
\centering
\includegraphics[width=0.5\textwidth]{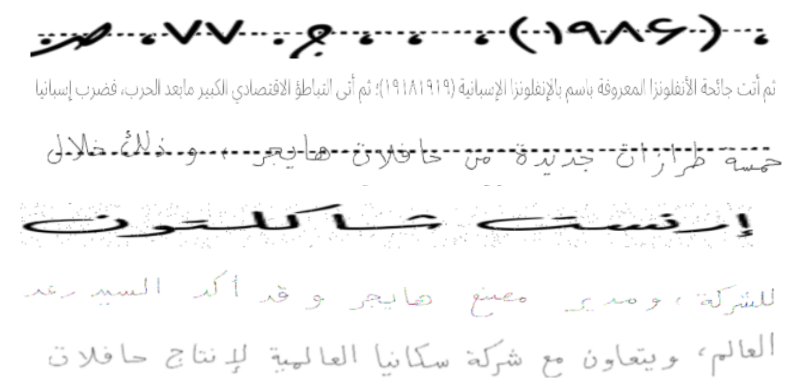}
\caption{Examples of Augmented Data: (a) lined paper, (b) Background Noise, (c) dotted paper, (d) Low-Resolution Effect, (e) Motion Blur.}
\label{fig:augmented_images}
\end{figure}

Regarding the level of augmentation applied, Figure~\ref{fig:augmented_images} shows some image examples after having applied various distortion types. Variation will definitely make the model learn real-world inconsistencies effectively and thus make the models robust in different handwriting and printed formats.

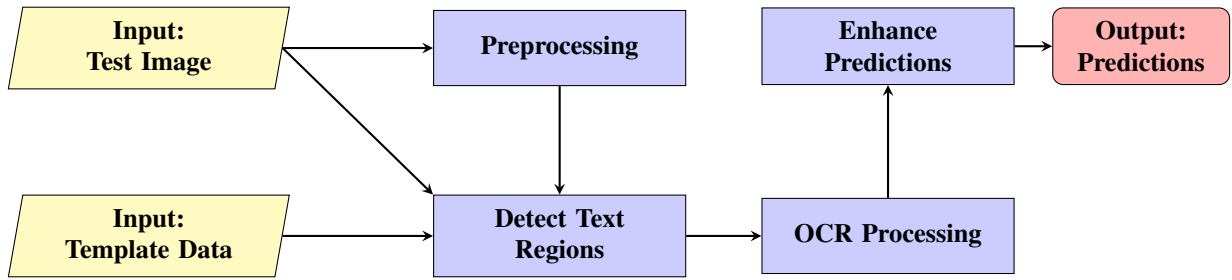
\begin{figure*}[!h]
    \centering
    \begin{tikzpicture}[node distance=1cm]
        \tikzset{
            startstop/.style = {rectangle, rounded corners, text width=2cm, minimum height=1cm, draw=black, fill=red!30, font=\bfseries, align=center, inner sep=5pt},
            process/.style = {rectangle, text width=3cm, minimum height=1cm, draw=black, fill=blue!20, font=\bfseries, align=center, inner sep=5pt},
            decision/.style = {diamond, text width=3cm, minimum height=1cm, draw=black, fill=green!30, font=\bfseries, align=center, inner sep=5pt},
            data/.style = {trapezium, trapezium left angle=80, trapezium right angle=100, text width=3cm, minimum height=1cm, draw=black, fill=yellow!30, font=\bfseries, align=center, inner sep=5pt},
            arrow/.style = {thick,->,>=stealth}
        }
        
        \node (inputImage) [data, yshift=1.5cm] {Input: \\ \textbf{Test Image}};
        \node (inputJSON) [data, yshift=-1cm] {Input: \textbf{Template Data}};
        
        \node (preprocess) [process, right=2cm of inputImage] {Preprocessing};
        \node (detect) [process, right=2cm of inputJSON] {Detect Text Regions};
        
        \node (ocr) [process, right=of detect] {OCR Processing};
        
        \node (enhance) [process, above=of ocr, yshift=0.5cm] {Enhance Predictions};
        \node (output) [startstop, right=of enhance, xshift=-0.5cm] {Output: \textbf{Predictions}};

        \draw [arrow] (inputImage) -- (preprocess);
        \draw [arrow] (inputImage.east) -- (detect.north west);
        \draw [arrow] (inputJSON) -- (detect);
        \draw [arrow] (preprocess.south) -- (detect.north);
        \draw [arrow] (detect) -- (ocr);
        \draw [arrow] (ocr.north) -_ (enhance.south);
        \draw [arrow] (enhance) -- (output);

    \end{tikzpicture}
    \caption{A visualization of the Main Module Pipeline starting with 2 inputs Test Image and Template Data. Applying the preprocessing Step to get the Preprocessed version of the test image then passing the original Test Image, Preprocessed Image, and Template Data to get the target text fields. The Third stage is the OCR process on all Text fields extracted. After all that Enhancing Prediction comes to function to minimize errors on given Pre-Defined nature of the text fields}
    \label{fig: Main Module Pipeline}
\end{figure*}

\subsection{Main Module}\label{main module}

The Main Module is the core engine that controls all other AI components. It handles the incoming image, performs all necessary operations that would be specified in the upcoming sections, and returns Predictions. The main module process can be visualized in Figure \ref{fig: Main Module Pipeline}. \\\\
\textbf{Template Data}:\\
The input Template Data contains all the information about the receipt or document that is predefined and used for Slicing input images and Enhancement of the prediction.
The template Data that contains predefined data can be clearfied as follows:  
\begin{itemize}
    \item shapes: An array of text fields in the document that contains
    \begin{itemize}
        \item "type": The type of receipt field used in the Enhancement stage or equipping the Detection Model in the process and it has 4 possible values:
        \begin{itemize}
            \item "Single Line": It means that the field text is an arbitrary line.
            \item "Multiple Lines": The text field is a paragraph.
            \item "Number": The text in the field contains only digits.
            \item "Date": The text is a date. 
            \item "Defined Label": The text is one of defined possibilities.
        \end{itemize}
        \item "points": It's the bounding boxes used for the extraction of text regions.
        \item "possibilities": A list of all possibilities of the existing text value (only applied when "type = Defined Label")
    \end{itemize}
    \item "imageData": The template image.
\end{itemize}

\textbf{Preprocessing}:\\
The preprocessing stage consists of 4 operations done on the input image.
\begin{enumerate}
    \item Converting the image into greyscale
    \item Applying the Fast Non-Local Means (FNLM) Denoising algorithm to effectively reduce noise and eliminate light lines and dots that may be present beneath the written content. The algorithm leverages the similarity between pixels in a non-local neighborhood, promoting noise removal while preserving image details. This approach can be particularly beneficial in cleaning up subtle artifacts without affecting the integrity of the underlying text or features. 
    
    Mathematically, the denoising process can be described by the following equation:
    
    \begin{equation}
        \hat{I}(x) = \frac{1}{C(x)} \sum_{y \in \Omega(x)} \mathcal{K}(x, y) I(y)
    \end{equation}
    
    where \( \hat{I}(x) \) is the denoised image, \( I(y) \) represents the intensity of pixel \( y \), \( \Omega(x) \) is the neighborhood around pixel \( x \), and \( \mathcal{K}(x, y) \) is a similarity function that determines the weight of pixel \( y \) relative to pixel \( x \). The normalization factor \( C(x) \) ensures that the weights sum to one.
    
    \item Binarizing the image using thresholding.
    \item Applying morphological transformations with a \( 3 \times 3 \) kernel, performing the Opening operation, which involves a sequence of erosion followed by dilation. This operation is specifically designed to remove small defects, such as isolated dots and noise, that may interfere with the Optical Character Recognition (OCR) accuracy. The opening operation helps in cleaning the image by eliminating tiny unwanted objects while preserving important structures, such as the characters. 

    The morphological opening operation can be expressed as:
    
    \begin{equation}
        \text{Open}(I) = \delta(\delta(I))
    \end{equation}
    
    where \( \delta(I) \) represents the erosion of the image \( I \), and the result of this erosion is then dilated to restore the size of the original objects.
    
\end{enumerate}
\textbf{Detect Text Region}:\\
This stage takes Template Data, Test Image, and Preprocessed image as input. The stage consists of several operations:
\begin{enumerate}
    \item Converting both the test image and the template image to grayscale.
    \item Extracting distinctive and scale-invariant key points from both grayscale images using the Scale-Invariant Feature Transform (SIFT) algorithm. The SIFT algorithm identifies key points in an image that are invariant to scale, rotation, and affine transformations, making it highly effective for tasks such as object recognition and image matching. These key points are characterized by their distinctive local image patterns, which can be reliably tracked across different views of the same scene. 
     
    \item Brute-Force Matching with L2 Norm as The Euclidean distance (\( L_2 \) Norm) is used to measure the similarity between descriptors. The formula for the Euclidean distance between two descriptors \( d_1 \) and \( d_2 \) of dimension \( n \) is given by:
    \begin{equation}
        D = \sqrt{\sum_{i=1}^{n} (d_{1i} - d_{2i})^2}
    \end{equation}
    
    For each descriptor in the first image, the algorithm finds the descriptor in the second image that has the smallest \( L_2 \) distance.
    
    Cross-Check for Mutual Validation to improve the accuracy of feature matching, Cross-Checking (mutual validation) is applied. This ensures that if a feature in Image A is matched to a feature in Image B, then the same feature in Image B should also match back to the same feature in Image A.
    
    \item 
    Computing the homography matrix \( H \) using the Direct Linear Transformation (DLT) algorithm or RANSAC-based estimation.
    
    Mathematically, the homography matrix \( H \) is a \( 3 \times 3 \) transformation matrix that maps points from one image to another:
    
    \begin{equation}
        \begin{bmatrix} x' \\ y' \\ w' \end{bmatrix} =
        H \begin{bmatrix} x \\ y \\ 1 \end{bmatrix}
    \end{equation}
    
    where \( H \) is defined as:
        
    \begin{equation}
        H =
        \begin{bmatrix} 
        h_{11} & h_{12} & h_{13} \\ 
        h_{21} & h_{22} & h_{23} \\ 
        h_{31} & h_{32} & h_{33} 
        \end{bmatrix}
    \end{equation}
    
    \item Utilizing the homography matrix to map the corresponding values from the template image onto the test image through perspective transformation.  
    \item Extracting the corresponding bounding boxes from the preprocessed image for further analysis.  

\end{enumerate}
\textbf{OCR Processing}:\\
This stage receives a series of text field images and utilizes detection and OCR recognition models. The stage consists of the following simple operations by for each of the input text fields:
\begin{enumerate}
    \item Check the type of the text field\\
    If the text field is defined as paragraph then apply the Detection Module to segment and extract images of existing lines in the paragraph before recognizing each line.\\
    Otherwise proceed to the Recognition Model assuming there only exists one line.
\end{enumerate}
\textbf{Enhance Predictions}:\\
This stage is only applied to non-arbitrary defined fields.
if type == Number:
\begin{itemize}
    \item Decline any character that is not considered a digit.
\end{itemize}
else if type == Date:
\begin{itemize}
    \item Decline any predictions that are not in the form of DD/MM/YYYY while taking into consideration the different forms of date listing.
\end{itemize}
else if type == Defined Label:
\begin{itemize}
    \item Selecting the element from the possibilities list that has the closest Levenshtein distance to the given target string. The Levenshtein distance, also known as the edit distance, is a measure of the minimum number of single-character edits (insertions, deletions, or substitutions) required to transform one string into another. By calculating the Levenshtein distance between the target string and each element in the Predefined Possibilities list, we can identify the element that is most similar to the target.

\end{itemize}

\section{Results and Discussion} \label{results section}
\subsubsection{Text Detection \& Line Segmentation}
The difference between the initial weights "best-weights of the DBNet++ (\hyperref[Liao2023]{Liao et al.,2023}) Model" and our Fine-Tuned weights on The Arabic dataset can be seen in Table \ref{table1}. The two models were tested on customized Arabic handwritten and non-handwritten data with precise annotations at the line level.
The evaluation of the model is using the following metrics: 
\begin{itemize}
    \item {Precision: measure how accurately the model segments the lines.}
    \item {Recall: measure how effectively the model captured most of the lines in the images.}
    \item {-F-Measure: balance performance in terms of both accuracy and completeness of the line segmentation.}
\end{itemize}

\begin{table}[htbp]
\caption{Detection Model performance}
\vspace{-10pt}
\begin{center}
\begin{tabular}{|c|c|c|c|}
\hline
\textbf{Model} & \textbf{Precision} & \textbf{Recall} & \textbf{F-Measure} \\
\hline
Our Detection Model & 81.66 & 78.82 & 79.07 \\
\hline

\end{tabular}
\label{table1}
\end{center}
\end{table}

\begin{figure}[htbp]
    \centering
    \vspace{-15pt}
    \begin{minipage}[b]{0.3\linewidth}
        \centering
        \includegraphics[width=\linewidth]{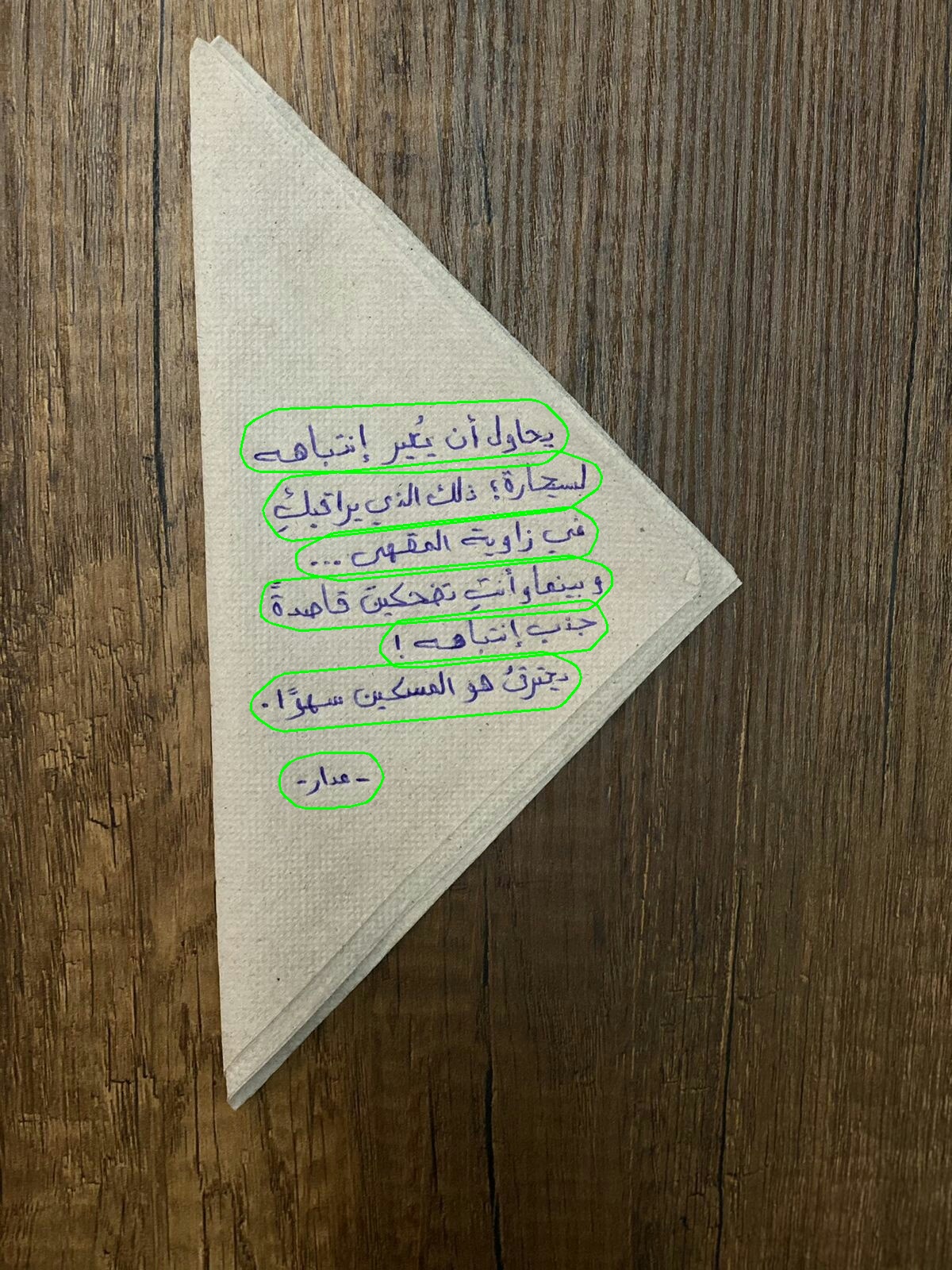}
        \label{fig:1-after}
    \end{minipage}
    \hspace{0.01\linewidth}
    \begin{minipage}[b]{0.41\linewidth}
        \centering
        \includegraphics[width=\linewidth]{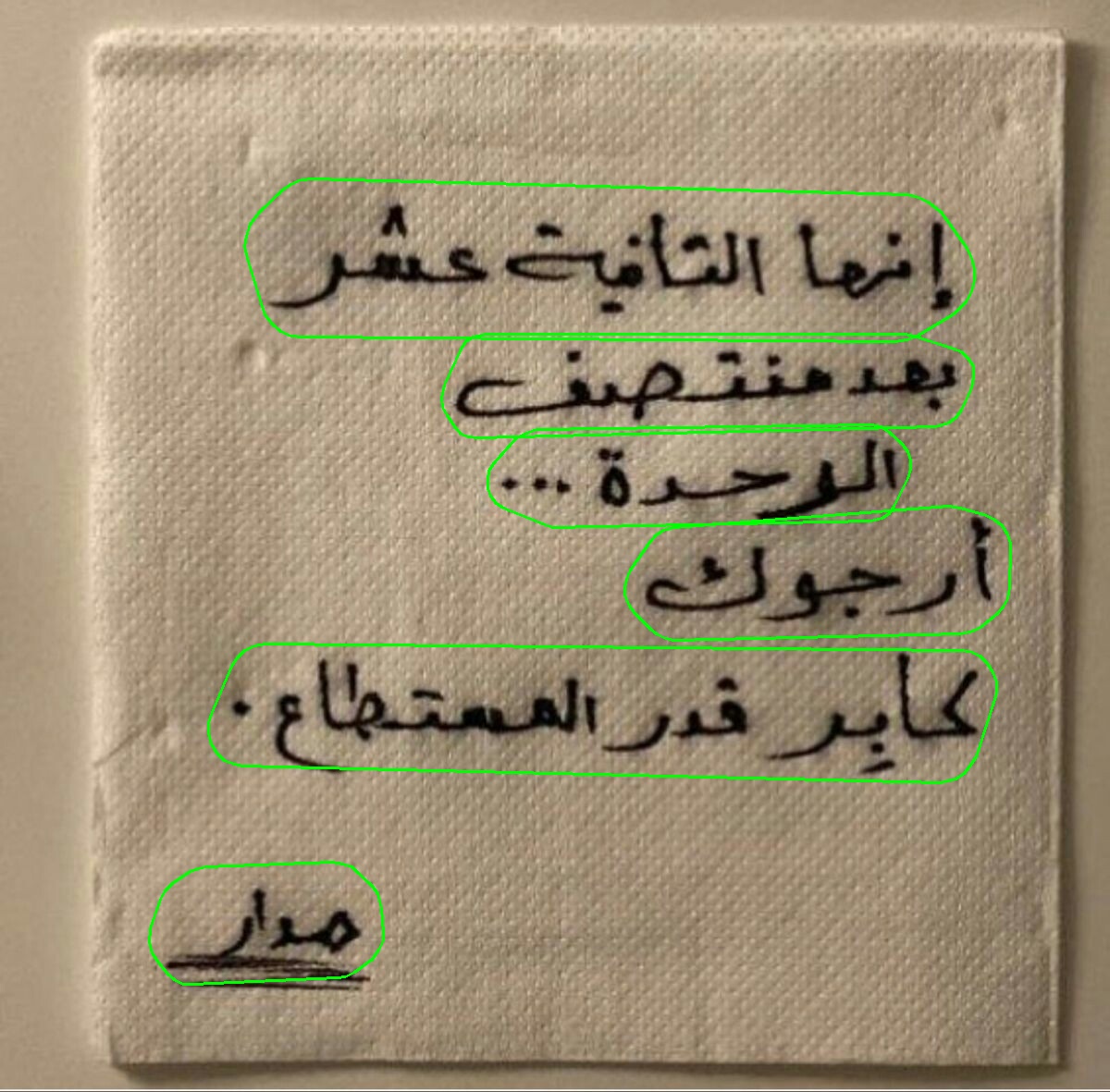}
        \label{fig:3-after}
    \end{minipage}
    \vspace{-15pt}
    \caption{Visualizing Line Segmentation Results}
    \vspace{-10pt}
\end{figure}

\subsection{Optical Character Recognition}
\subsubsection{Evaluation Metrics}
The following metrics serve as the key to assess the performance of our OCR model:
\begin{itemize}
 \item \textbf{Character Error Rate (CER):} calculates the incorrect recognition of characters in a percentage.
\item \textbf{Word Error Rate (WER) :} Measures the percentage of incorrectly recognized words.
\end{itemize}

\subsubsection{Results on Printed Text (Without Augmentation)}
Table~\ref{tab:printed_results} presents the performance of our model on printed Arabic text. The model achieves a low CER and WER, demonstrating its high accuracy in recognizing structured printed text.
\begin{table}[ht]
\centering
\caption{CER and WER for Printed Text (Without Augmentation)}
\label{tab:printed_results}
\begin{tabular}{|c|c|}
\hline
\textbf{Metric} & \textbf{Score} \\
\hline
CER & 0.59\% \\
WER & 1.72\% \\
\hline
\end{tabular}
\end{table}

\subsubsection{Results on Online-KHATT Dataset (Without Augmentation)}

We also evaluate our approach on the benchmark of the Arabic Online-KHATT dataset encompassing various ways of Arabic writing. Our system substantially reduces state-of-the-art error rates developed in \hyperref[alwajih2022]{Alwajih et al., 2022} in high margins based on both of CER and WER mentioned in Table~\ref{tab:khatt_comparison}.
\begin{table}[ht]
\centering
\caption{Comparison of CER and WER on Online-KHATT Dataset (Previous work from \hyperref[alwajih2022]{Alwajih et al., 2022}}
\label{tab:khatt_comparison}
\begin{tabular}{|c|c|c|}
\hline
\textbf{Metric} & \textbf{Previous Work} & \textbf{Our Model} \\
\hline
CER & 22.00\% & \textbf{7.91\%} \\
WER & 56.00\% & \textbf{31.41\%} \\
\hline
\end{tabular}
\end{table}

W\subsubsection{Results on Printed and Handwritten Text (With Augmentation)}
To enhance robustness, our model is trained with augmented data that includes background noise, motion blur, and resolution degradation. After the implementation of such augmentation, results are given by CER/WER values as depicted in Table~\ref{tab:augmented_results}.
\begin{table}[ht]
\centering
\caption{CER and WER for Printed and Handwritten Text (With Augmentation)}
\label{tab:augmented_results}
\begin{tabular}{|c|c|c|}
\hline
\textbf{Metric} & \textbf{Printed Text} & \textbf{Handwritten Text} \\
\hline
CER & 0.93\% & 8.19\% \\
WER & 2.20\% & 32.42\% \\
\hline
\end{tabular}
\end{table}

A slight increase in error rates is expected due to the complexity introduced by augmentations, making the model more resilient to real-world distortions.

\begin{figure}[!hbtp]
    \centering
    \includegraphics[width=0.999\linewidth]{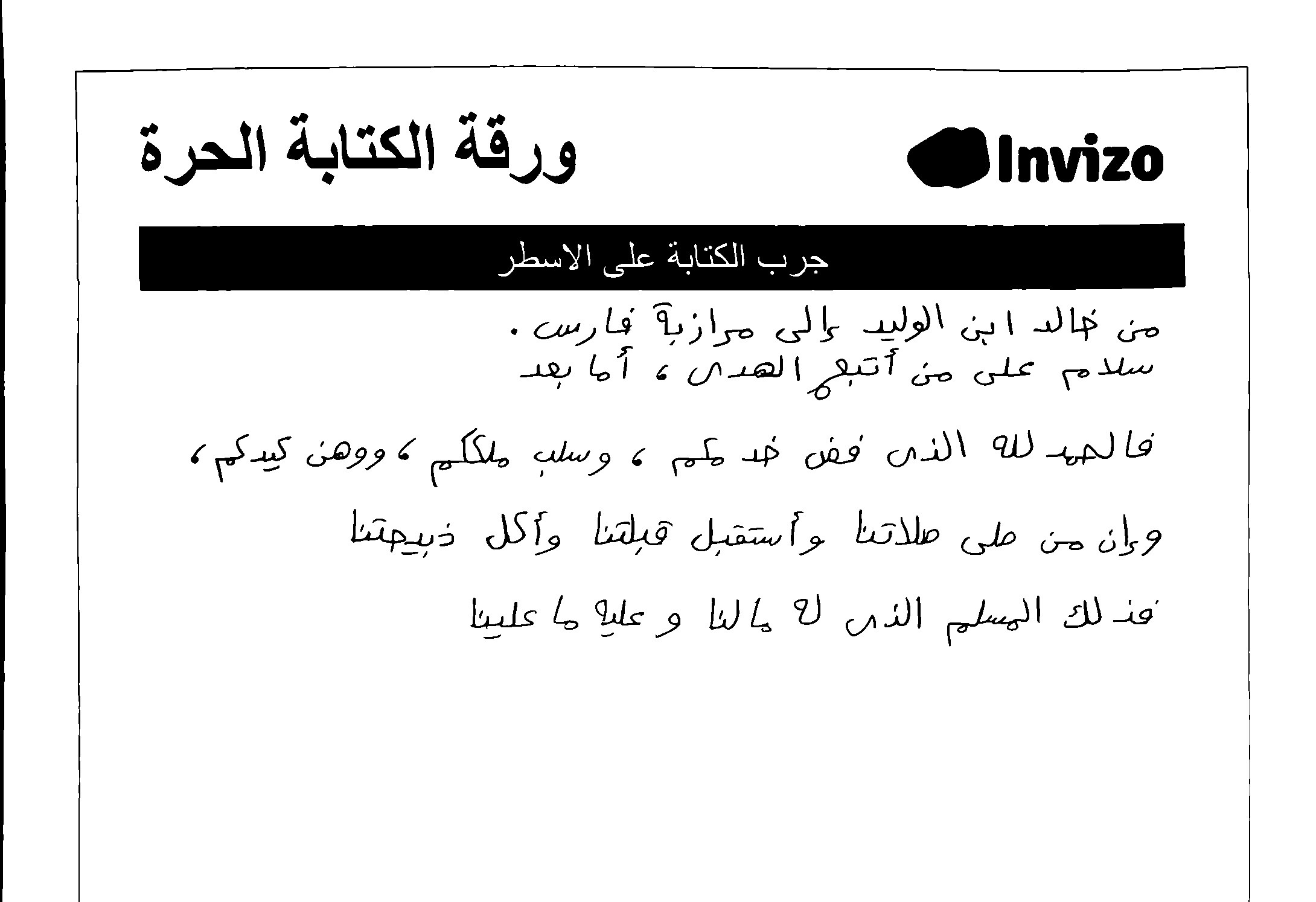}
    \caption{Visualization of the Denoising after Binarization}
    \label{fig: Visualization of the Denoising after Binarization}
\vspace{-10pt}
\end{figure}
\subsubsection{Pre-Processing}
The Pre-Processing aims to clean the image of lines from defects while maintaining clean writing without affecting the text shape. The tests contained multiple types of documents with several empty text field shapes.\\
First, the Denoising via Fast Non-Local Means Denoising, performed very well in terms of removal of noise but we also chose that method to remove the line underneath the writings. So in cases there is a light straight line or dashed line the operation was successful, as show in Figure \ref{fig: Visualization of the Denoising after Binarization} after the binarization operation.

\begin{figure}[!hbtp]
    \centering
    \includegraphics[width=0.999\linewidth]{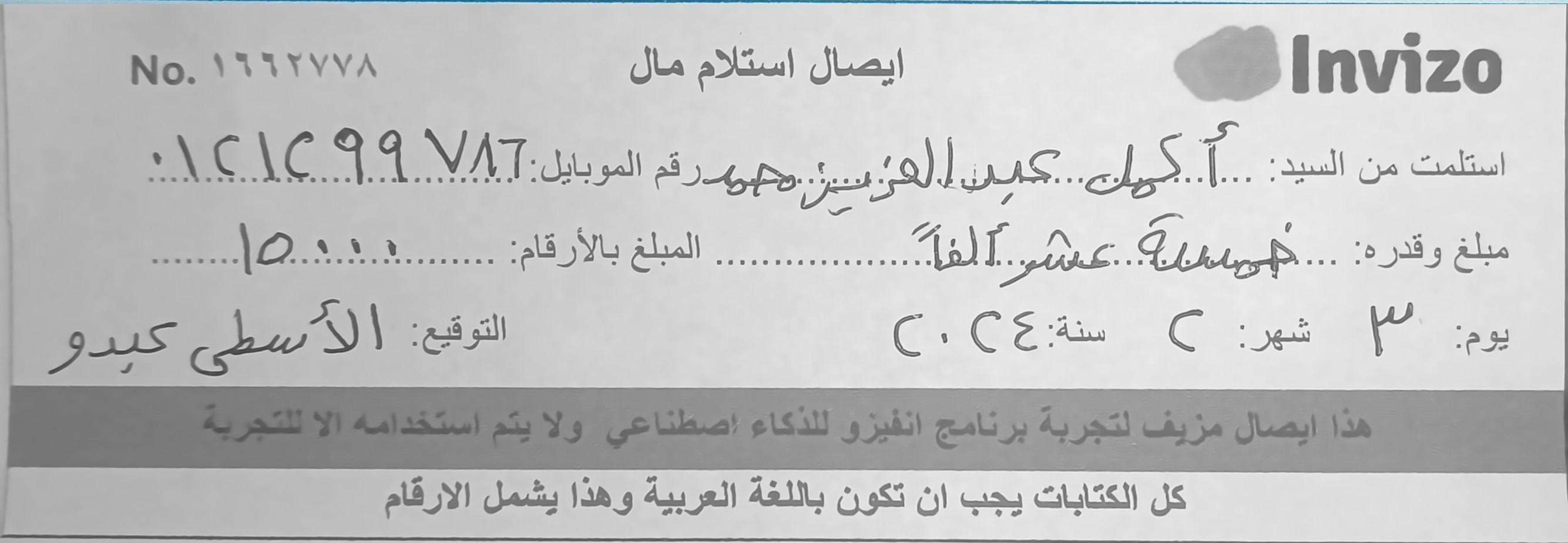}
    \caption{Visualization of the Denoising on darker inked line}
    \label{fig: Visualization of the Denosising on darker inked line}
\vspace{-7pt}
\end{figure}
Yet the operation fails to clear lines with more darker ink as shown in Figure \ref{fig: Visualization of the Denosising on darker inked line}. We tried different approaches to remove the dotted line via detection and removal, yet it ended up by affecting the other clean samples and sometimes the writings.

\begin{figure}[!hbtp]
    \centering
    \includegraphics[width=0.999\linewidth]{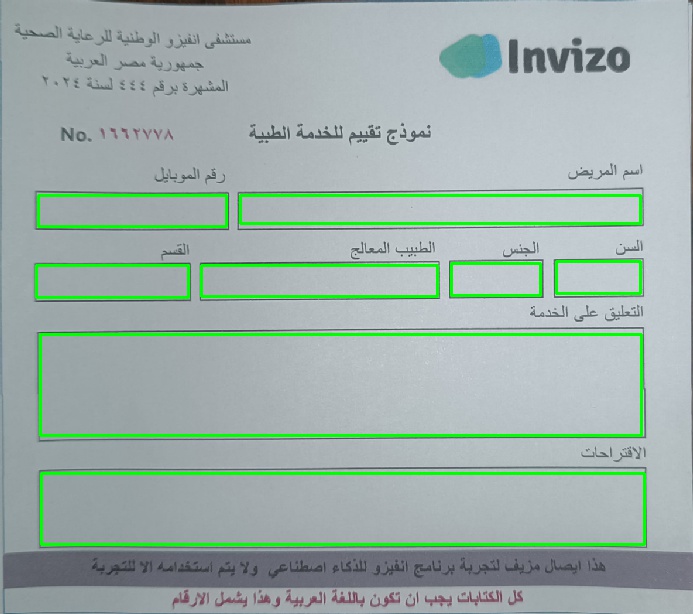}
    \caption{Visualization of the Template Image with Annotations}
    \label{fig: Template Image with Annotations}
\vspace{-7pt}
\end{figure}
\subsubsection{Detection of Text Region}
\begin{figure}
    \centering
    \includegraphics[width=0.999\linewidth]{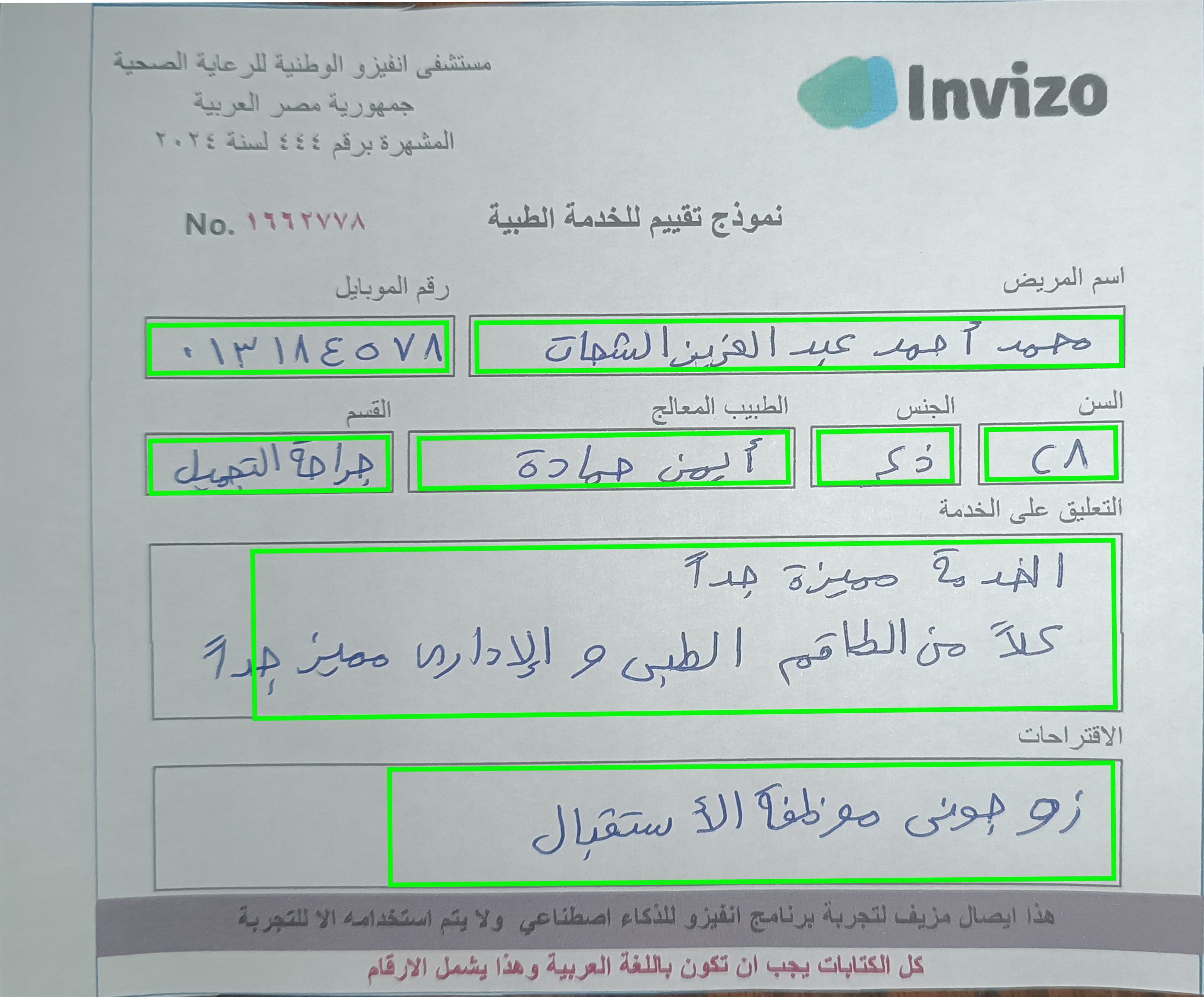}
    \caption{Visualization of a Test Image with Annotations}
    \label{fig: Test Image with Annotations}
\vspace{-7pt}
\end{figure}

\begin{figure}
    \centering
    \includegraphics[width=0.999\linewidth]{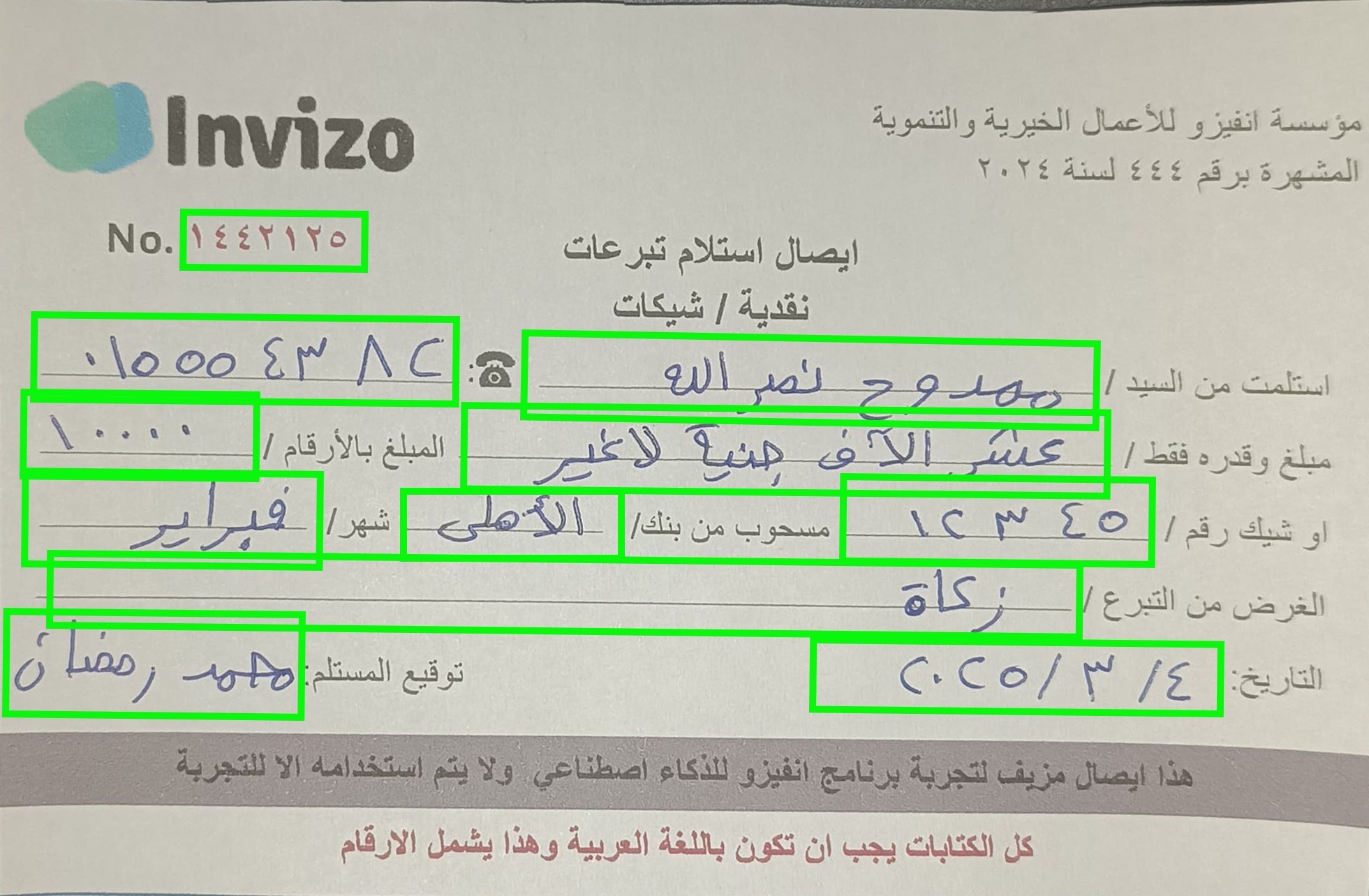}
    \caption{Visualization of a Test Image with Annotations}
    \label{fig: Another Test Image 2 with Good Annotations}
\end{figure}

In that process of fields extraction based on Pre-Defined text fields on another image of the same document type. The template image is defined as the Figure \ref{fig: Template Image with Annotations}, while the test image with annotations produced by the template image is in Figure \ref{fig: Test Image with Annotations}. 
\\
Some results as seen have perfect match. Others seem not to fit the target very well and that is due to the image's lack of details in the document type which produces fewer features to be computed against each other for perfect matching. However, the results on most of the documents and receipts are very satisfactory, as the ones found in Figure \ref{fig: Another Test Image 2 with Good Annotations}.

\section{Conclusion} \label{conclusion section}
The development of a full Arabic OCR system that can handle diverse real-life receipts means big steps toward document digitization. This is where the gap between basic research and industrial practice has been successfully bridged, offering solutions for the complex nature of the Arabic script with the challenges imposed by heterogeneous document formats. A proposed system that focuses on robustness, accuracy, and adaptability tries to overcome most of the critical limitations in existing approaches. The large variation in types of receipts such as retail invoices, restaurant bills, and official certificates makes the system very effective.\\
That is highly beneficial and is considered the major strength of the system: its capability to handle diversified document formats, enabled through advanced text detection combined with robust OCR recognition. The detection module plays a significant role in the system by precisely localizing text regions even on cluttered or noisy receipts. This ensures that key information, such as totals, dates, and signatures, can be reliably extracted. It is designed to be flexible with the variation in layout and content, therefore very versatile for handling documents with different structures.\\
The detection of the Text comes either from Feature Matching given a pre-defined document that we know where the text shall be or the detection model that proved its effectiveness in Line Segmentation in paragraph text.

The OCR module, combining a CNN-based feature extractor with a transformer-based sequence model, achieves significant improvements in accuracy over conventional methods. Training on a composite dataset spanning handwritten text, synthetic printed content, and numerals ensures generalization across text styles and noise conditions. Innovations such as \textbf{sinusoidal positional encoding} and \textbf{multi-head attention mechanisms} enhance the model’s ability to handle Arabic script’s cursive dependencies, resolving ambiguities in connected letters and overlapping strokes.\\
It involved data augmentation to make sure the model generalizes well for most real-world conditions, including background noise, motion blur, low resolution, and distortions to simulate most document imperfections that could be encountered in practical use. This augmentation further makes sure that the model will work reliably even when it is fed with more challenging inputs, like receipts scanned of variable quality and noise. It goes on to prove the capability of dealing with document variability and being deployed into the real world for industries like retail, finance, or government services where automation in document processing may reduce manual interference (human errors) and increase efficiency. However, the main part of the OCR engine needs further improvement. As it can perform well on a handwriting style and performs very badly on other styles. To be more specific the limitation is the handwriting styles that are basic or more like the regular printed Arabic are less likely to have any problems in the recognition. On the other hand, more styles are complex and has a unique cursive nature which the Recognition Model has trouble recognizing.

In conclusion, the Arabic OCR system proposed here was a big step toward the automation of document processing for Arabic text. By solving the challenges of handwritten and printed Arabic text, the system introduced here will achieve both high accuracy and practical utility by employing a hybrid architecture together with advanced detection and augmentation techniques. However, the limitation of the system aligns with the recognition module. This work contributes to the advance of Arabic OCR technology and provides a reliable solution for real-world document digitization tasks.

\section{Future Work} \label{future work section}
While the current system's performance is promising, key areas require further enhancement to ensure robustness and scalability in real-world applications.\\
First is the preprocessing of \textbf{real world images}, despite the implementation of preprocessing techniques and data augmentation, still poses a challenge. Most of them have inconsistent backgrounds and low resolution with noise, thus affecting the model's performance in terms of the capability of making correct predictions.To address these issues, integrating super-resolution models such as ESRGAN or SRCNN could significantly improve the quality of low-resolution images before feeding them into the model. This would enhance the visibility of fine details that may be blurry or not well-defined, which is expected to improve recognition accuracy.\\
Another limitation relates to the \textbf{size and diversity of the dataset} used for training. The present dataset (Online-KHATT) is useful but not diverse enough concerning variations in handwriting, complex backgrounds, and real-world conditions. Due to this, the model performance suffers concerning generalization. To handle this, more data collection needs to be done to provide a wide variation of handwritten text with more diversity in augmentation, like lighting conditions, text alignment, and orientations. The increase in the dataset will help the model learn from a wider variety of handwriting styles and environments, enhancing its ability to handle documents coming from various sources. In that context, we intend to expand our Naqsh Arabi OCR dataset to be a massive dataset with various writing styles.\\
A further significant improvement can be made by \textbf{continuous learning}. Currently, the model is trained on a fixed dataset, and after training, it becomes static. However, in real-world applications, new data keeps appearing, such as new handwriting styles or different document formats. This may lead to degradation in performance over time since the model never sees these new examples. This model learns with the introduction of new data without requiring a full retraining process with the use of continuous learning. Utilizing either of the two most popular techniques (incremental learning) or (online learning), it continuously updates the parameters and learns fresh examples on the go. This will be able to keep the system updated with evolving handwriting styles and document layouts, among other real-world variations that may not have been present in the initial training dataset. Continuous learning would maintain accuracy with more recent data and reduce the time and cost required for retraining. This would make a model increasingly robust, better at generalizing to new unseen data, and current in the ever-evolving practical world.


%


\ifCLASSOPTIONcaptionsoff
  \newpage
\fi

\appendices
\section{Naqsh Arabi Dataset Content Retrieval Wikipedia Pages} \label{Naqsh Dataset Appendix}
The following is a list of the pages used for the extraction of the Arabic Text data categorized by the genre. Each of the following lacks "https://ar.wikipedia.org/wiki/" before it to form URL and replace regular space with "\_".
\vspace{-10pt}
\subsection{Geography}
\scriptsize {\foreignlanguage{arabic}{
جغرافيا, خريطة, قارة آسيا, قارة أفريقيا, قارة أمريكا الشمالية, قارة أمريكا الجنوبية, قارة أنتاركتيكا, قارة أوروبا, قارة أستراليا, المحيط الهادئ, المحيط الأطلسي, المحيط الهندي, بحر القرم, بحر الآرال, نهر النيل, نهر الأمازون, نهر الدانوب, نهر الميسيسيبي, نهر اليانغتسي, الصحارى, صحراء التهامة, صحراء الصحراء العظمى, صحراء الكلورادو, صحراء ساهارا, صحراء غابات الأمازون, جبال الهيمالايا, جبال الألب, جبال الأنديز, جبال روكي, جبال الطور, غابات الأمطار, الغابات الاستوائية, المنطقة القطبية, التضاريس, المناخ, المناطق المعتدلة, المناطق الاستوائية, القطب الشمالي, القطب الجنوبي, بحيرة الحدود, الحدود البحرية, الغلاف الجوي, الظواهر الطبيعية, الرياح, الزلازل, البراكين, المياه الجوفية, السواحل, الجزيرة, خليج العرب, البحر الأبيض المتوسط, بحر الشمال, بحر العرب, مضيق هرمز, مضيق بوسفور, مضيق الدردنيل, مضيق مالاكا, المحيط القطبي, الخلجان, توزيع الموارد الطبيعية, المعادن, النفط والغاز, الحدائق الوطني, حوض النهر, السدود, الأنهار الكبرى, الصيد البحري, الإحياء الطبيعي, الحيوانات البرية, التنوع البيولوجي, الجبال الزرقاء, التلوث البيئي, التغيرات المناخية, الاحتباس الحراري, الفيضانات, الجفاف, البر الاستوائي, البر الصحراوي, البيئة البحرية, المحيطات الهنغارية, البحيرات الاستوائية, البحيرات الحوضية, السواحل المدارية, المعادن الطبيعية, الفضاء, الحدائق الاستوائية, الموارد الطبيعية في أفريقيا, الأنظمة البيئية, العوامل المناخية, الزراعة الاستوائية, الزراعة الصحراوية, الموارد المائية, التربة, الترشيح, التنوع البيئي البري, المياه المالحة, التضاريس الطبيعية, دورة المياه, التوازن البيئي, الحرائق الغابات, النباتات الاستوائية, النباتات الصحراوية, الأنظمة المائية, البيئة الاستوائية, الغابات المدارية, الجبال الاستوائية, الحدود الجغرافية, الطرق البحرية, الأنظمة الجغرافية, الدور البيئي
}
}
\vspace{-10pt}
\subsection{History}
\scriptsize {\foreignlanguage{arabic}{
تاريخ, تاريخ العرب, تاريخ العالم, تاريخ الشرق الأوسط, تاريخ مصر, تاريخ العراق, تاريخ فلسطين, تاريخ الشام, تاريخ المغرب العربي, تاريخ اليمن, تاريخ السعودية, تاريخ تونس, تاريخ الجزائر, تاريخ ليبيا, تاريخ المملكة المتحدة, تاريخ فرنسا, تاريخ إسبانيا, تاريخ إيطاليا, تاريخ ألمانيا, تاريخ روسيا, تاريخ الولايات المتحدة, تاريخ الصين, تاريخ اليابان, تاريخ الهند, تاريخ مصر القديمة, تاريخ الفرس, تاريخ اليونان, تاريخ روما, تاريخ العصر الحديث, تاريخ العصور الوسطى, تاريخ الديانات, تاريخ الإسلام, تاريخ الكنيسة الكاثوليكية, تاريخ الديانة اليهودية, تاريخ المسيحية, تاريخ الديانات السماوية, تاريخ الفراعنة, تاريخ الحضارة الفرعونية, تاريخ الهند القديم, تاريخ الصين القديم, تاريخ المدن التاريخية, تاريخ المجتمعات القديمة, تاريخ الاقتصاد, تاريخ الفنون, تاريخ العمارة, تاريخ العلوم, تاريخ التعليم, تاريخ المرأة, تاريخ الرياضة, تاريخ الطب, تاريخ الطب في العصور القديمة, تاريخ الفيزياء, تاريخ الكيمياء, تاريخ الرياضيات, تاريخ الفلسفة, تاريخ الأدب, تاريخ الشعر, تاريخ المسرح, تاريخ السينما, تاريخ الفن الإسلامي, تاريخ الخط العربي, تاريخ الحضارة الإغريقية, تاريخ الحضارة الرومانية, تاريخ الحضارة البيزنطية, تاريخ الحضارة الفرعونية, تاريخ الحضارة السومرية, تاريخ الحضارة المصرية, تاريخ الحضارة الفارسية, تاريخ الحضارة الإسلامية, تاريخ الغزو الإسلامي, تاريخ الفتح الإسلامي, تاريخ الإمبراطورية العثمانية, تاريخ الإمبراطورية الرومانية, تاريخ الإمبراطورية الفارسية, تاريخ الحروب العالمية, تاريخ الحرب العالمية الأولى, تاريخ الحرب العالمية الثانية, تاريخ الاستعمار, تاريخ الثورة الفرنسية, تاريخ الثورات الأمريكية, تاريخ الثورة الروسية, تاريخ الاقتصاد العالمي, تاريخ الاستقلال, تاريخ حقوق الإنسان, تاريخ الديمقراطية, تاريخ الاشتراكية, تاريخ الرأسمالية, تاريخ الأنظمة السياسية, تاريخ التطورات السياسية, تاريخ الأسلحة, تاريخ الحروب والصراعات, تاريخ الحروب الإسلامية, تاريخ الحروب الأهلية, تاريخ الانتفاضات, تاريخ الاحتلال الفرنسي, تاريخ الاحتلال الإنجليزي, تاريخ الاحتلال الإيطالي, تاريخ الاحتلال البرتغالي, تاريخ الاحتلال العسكري, تاريخ الاستعمار الإنجليزي, تاريخ الاستعمار الفرنسي, تاريخ الاستعمار البرتغالي, تاريخ الاستعمار الإيطالي, تاريخ الاحتلال الاستعماري, تاريخ التدخلات العسكرية, تاريخ التوسع الإمبراطوري, تاريخ الاستقلال الوطني, تاريخ الثورات العربية, تاريخ الصحافة العربية, تاريخ المثقفين العرب, تاريخ الأدب العربي, تاريخ العلوم العربية, تاريخ الجامعة العربية, تاريخ الزعامات العربية, تاريخ الثورات العالمية 
}
}
\vspace{-10pt}
\subsection{Information Technology}
\scriptsize {\foreignlanguage{arabic}{
تكنولوجيا المعلومات, أنظمة المعلومات, تقنية, معلومات تقنية, حاسوب, برمجة, ذكاء اصطناعي, تعلم آلي, شبكة حاسوب, إنترنت, تقنية الجيل الرابع, تقنية الجيل الخامس, إلكترونيات, برمجيات, أنظمة تشغيل, خوارزميات, خوادم, حاسوب محمول, دور المبرمج, تطوير البرمجيات, شبكات الاتصال, قواعد البيانات, إدارة المشروعات التقنية, تكنولوجيا البرمجيات, الأنترنت الأشياء, أمن المعلومات, تشفير, المعلومات السرية, دورة الحياة البرمجية, الذكاء الاصطناعي التطبيقي, منهجية البرمجة, التطبيقات الهاتفية, تطبيقات الويب, تكنولوجيا الحوسبة, مطور البرمجيات, الذكاء الاصطناعي العميق, تقنيات الواقع المعزز, تقنيات الواقع الافتراضي, تقنيات المؤتمرات عن بعد, معالجة اللغة الطبيعية, أدوات التطوير, نظم الاستشعار, تصميم الأنظمة, البحث الإنترنتي, تخزين البيانات, تحليل البيانات, تخزين سحابي, الذكاء الصناعي, الحوسبة السحابية, تكنولوجيا الحوسبة الكمومية, البرمجيات الحرّة, مواقع الويب, خدمات الويب, التكنولوجيا الحديثة, البرمجيات المفتوحة المصدر, البحث في الويب, الروبوتات, تكنولوجيا الروبوتات, تطبيقات الذكاء الاصطناعي, الحوسبة المرنة, البرمجة الموجهة بالكائنات, البرمجة الوظيفية, البرمجة الإجرائية, إدارة المحتوى, النظم المتوازية, التعليم الالكتروني, المحتوى الرقمي, تطوير الويب, برمجة الويب, البحث الاكاديمي, التفاعل البشري الحاسبي, نظم التوصية, الذكاء الاصطناعي العام, الخوارزميات العددية, أنظمة الذكاء الاصطناعي, تصميم الواجهات المستخدم, مطور الويب, أدوات التحليل البياني, التطبيقات الرقمية, مكتبات البرمجيات, الحوسبة التي تديرها البرمجيات, إدارة البرمجيات, أتمتة البرمجيات, أدوات الأتمتة, الويب الدلالي, تقنيات التعلم العميق, تقنيات التعلم المشترك, الإنترنت الآمن, مقدمة إلى الذكاء الاصطناعي, إدارة البيانات, مستقبل البرمجة, أنظمة التشغيل الموزعة, الخوادم الموزعة, تحليل البيانات الكبيرة, أدوات تحليل البيانات, الأنظمة المعتمدة على الذكاء الاصطناعي, إدارة البنية الأساسية لتكنولوجيا المعلومات, نظم المعلومات الإدارية, دورة الحياة لتطوير البرمجيات, تقنيات الويب الجديدة, مستقبل التكنولوجيا, التقنيات الحديثة, أجهزة الكمبيوتر المحمولة, شبكات الإنترنت, الأنظمة الموزعة, النقل البيانات, الويب الاجتماعي, تكنولوجيا الاتصالات, تكنولوجيا الشبكات الاجتماعية, تكنولوجيا التجارة الإلكترونية, تقنيات التعليم الإلكتروني
}
}
\vspace{-10pt}
\subsection{Poetry}
\scriptsize {\foreignlanguage{arabic}{
شعر, أدب, رواية, قصة, شعر عربي, شعر حديث, شعر مكتوب, شعر مترجم, شعر صوفي, شعر غنائي, شعر فصيح, شعر عامي, قصيدة, قصيدة شعرية, شعر نبطي, شعر ملحمي, شعر عاطفي, شعر هجاء, قصص الأطفال, روايات عربية, الرواية الحديثة, قصة قصيرة, قصص تراثية, حكاية, سرد, أسلوب الرواية, أدب الطفولة, شعر تقليدي, شعر سوري, شعر مغربي, شعر مؤثر, شعر عربي مترجم, أدب مغاربي, قصيدة نثرية, شعر شعبي, شعر جماعي, شعر جديد, شعر تجريبي, قصص كلاسيكية, شعر وطنية, قصص حب, شعر تراثي, شعر رمزي, شعر تسجيلي, قصة عربية, قصص خيالية, قصة تاريخية, قصص بوليسية, أدب كلاسيكي, قصص خيالية علمية, قصص تراثية عربية, أدب رومانسي, قصص دينية, قصص اجتماعية, أدب الطفولة, شعر فلسفي, شعر إسلامي, شعر صوفي عربي, شعر فارسي, شعر تركي, شعر أوروبي, أدب فرنسي, أدب إنجليزي, شعر إنجليزي, قصص إنجليزية, قصص روسية, قصص صينية, قصص شعبية, قصص الأبطال, قصص طويلة, أدب تاريخي, قصص مغامرات, قصص فكاهية, شعر فارسي مترجم, أدب عربي حديث, شعر هندي, شعر إفريقي, شعر إنجليزي مترجم, قصص خيالية فانتازيا, شعر ما بعد الحداثة, رواية جديدة, قصة سودانية, قصص أطفال عربية, شعر حر, قصص تراثية مغربية, أدب أمريكي, أدب مصري, قصص غرامية, قصص معاصرة, قصص مغامرات عربية, أدب فلسطيني, قصص ألمانية, قصص مترجمة, أدب شعبي, شعر مترجم إلى العربية, أدب المخيلة, شعر مستقبلي, قصص تراثية قديمة, شعر حساس, شعر عالمي, قصة الأطفال العربية, قصص شعبية مغربية, قصص شعبية مصرية, قصص إيطالية, قصص إسبانية, قصص تركية, أدب مغربي, شعر إفريقي مترجم, شعر أمريكي, شعر عالمي حديث, شعر غربي, قصص إيرانية, أدب إيطالي, أدب إسباني, أدب تركي, أدب ألماني, أدب بلجيكي, أدب سويدي, قصص بلجيكية, قصص سويدية, أدب روسي, قصص فرنسية, قصص أمريكية
}
}
\vspace{-10pt}
\subsection{Mathematics}
\scriptsize {\foreignlanguage{arabic}{
رياضيات, جبر, تفاضل وتكامل, نظرية الأعداد, هندسة, إحصاء, تحليل رياضي, نظرية المجموعات, نظرية الألعاب, هندسة إقليدية, جبر خطِي, منطق رياضي, دوال, هندسة مستوية, حساب تفاضلي, حساب تكاملي, نظرية الاحتمالات, أنظمة خطية, معادلات تفاضلية, الرياضيات العددية, الهندسة التفاضلية, نظرية المصفوفات, نظرية الأنظمة, نظرية الشبكات, الرياضيات المالية, نظرية المقارنة, توافقيات, حساب متعدد المتغيرات, الرياضيات الحديثة, رياضيَّات مُلَتَحَمة, نظرية الجماعات, نظرية المجموعات المنفصلة, نظرية التعددية, خوارزميات, أرقام عشوائية, تحليل توافقي, مجموعات محددة, مصفوفات, نظريات العد, نظرية العلاقات, الحساب العددي, علم المعادلات الجبرية, حساب مصفوفات, التفاضل المركب, دوال متعددة المتغيرات, الرياضيات المنطقية, مصفوفات مربعة, الهندسة الرياضية, الرياضيات الفيزيائية, نظرية الربط, جبر تقليدي, أنظمة عشوائية, رياضيات معاصرة, مجموعة محدودة, شعاعيات, معدل تدفق الهواء, الدوال الخطية, نظرية الاحتساب, الرياضيات الخطية, مصفوفات مستقطبة, الرياضيات الذرية, التكامل العددية, التكامل العددي, نظرية الاحتمالات الشرطية, العلاقات المنطقية, التفاضل والتكامل, نظرية المعادلات الجبرية, الرياضيات الرقمية, علم الهندسة, النظرية الإحصائية, المعادلات الخطية, الرياضيات الهندسية, علم الرياضيات الحاسوبية, علم التكرار, نظرية الخطوط المنحنية, علم الهندسة الرياضية, جبر الخطوط, نظرية المعادلات غير الخطية, نظرية المقاييس, الهندسة التحليلية, الرياضيات التجريبية, نظرية المعادلات المعقدة, الرياضيات الفراغية, المعادلات الفائقة, النظرية التناسبية, نظرية الأرقام الصحيحة, النظرية البيانية, تحليل رياضي عددي, تحليل جبر متعدد المتغيرات, حساب المنطق, نظريات الهندسة الرياضية, رياضيات إحصائية, رياضيات مفاهيمية, رياضيات مؤسساتية, دوال غير خطية, المعادلات الديكارتية, مصفوفات الدوال, مجموعات أبسط, نظرية القياس العددي, هندسة المصفوفات, دوال رياضية, نظرية المعادلات التجريبية, رياضيات ثابتة, العلاقات التقريبية, نظرية المنطق العددي, تسلسل دالات, نظرية المجموعات الهامشية, المعادلات الجزئية, إستراتيجيات الإستدلال, المنطق التطبيقي, الرياضيات التاريخية, المعادلات المصفوفية
}
}
\vspace{-10pt}
\subsection{Languages}
\scriptsize {\foreignlanguage{arabic}{
علم اللغويات, لغويات نظرية, علم النحو, صوتيات, مفردات, علم الدلالة, نحو, دراسة اللغة, تركيب لغوي, علم التواصل, اللغويات العرفية, نحو عربي, لغويات تطبيقية, اللغويات السوسيولوجية, دلالات اللغة, اللغويات النفسية, مقارنة اللغات, علم اللغات التاريخية, علم اللغات التطبيقي, لغات البرمجة, دراسات اللغويات, ترجمة, ترجمة فورية, دراسة اللغات العالمية, اللغويات الاجتماعية, الدلالات المعنوية, نحو تطبيقي, اللغة والأدب, فهم اللغة, اللغة المنطوقة, اللغات الأوروبية, اللغات العربية, اللغات الآسيوية, اللغة الإنجليزية, اللغة الفرنسية, اللغة الألمانية, اللغة الروسية, اللغة الصينية, اللغة الإسبانية, اللغة الإيطالية, اللغة اليابانية, اللغة العبرية, اللغة التركية, اللغة الهندية, علم اللغات, دراسة اللسانيات, اللغويات الوظيفية, اللغويات البنائية, اللغويات التوليدية, علم النطق, اللغة الشعبية, اللغة الفصحى, اللغة العامية, التعابير اللغوية, اللغة والثقافة, اللغة والذاكرة, السيميائية, نحو التطوير, إعداد المترجم, اللغات الأفريقية, اللغات الآرية, اللغات السامية, اللغات المنغولية, اللغات المالايو-بوليغينية, اللغات النيجير-كونغو, اللغات الأمريكية الهنود, اللغات البولينيزية, اللغات الإسكندنافية, اللغات الرومانسية, اللغات الجرمانية, اللغات السلتيكية, اللغات الفنلندية, اللغات البربرية, اللغات الآبائية, اللغات الطويلة, اللغات الخفيفة, اللغات الصوتية, اللغات المستحدثة, اللغات الصينية الحديثة, اللغات الأمريكية الحديثة, اللغات المنغولية الحديثة, اللغات الإنجليزية الحديثة, اللغات التقليدية, اللغات المندثرة, تطور اللغات, مقارنة اللغات, اللغات المكتوبة, اللغات الشفوية, اللغات الوسيطة, تكنولوجيا اللغات, الترجمة الآلية, اللغويات الحاسوبية, الذكاء الاصطناعي واللغة, دراسة اللغات بالتعلم الآلي, تطبيقات اللغويات, لغويات تخطيطية, لغويات مقارنة, مفهوم اللغة, اللغات البرمجية, علم اللغات التطبيقي, نظام التعرف على الكلام, الترجمة الحاسوبية, الترجمة التقليدية, الترجمة التلفزيونية, النحو التقليدي, تعليم اللغات, دراسة اللهجات, دراسة الأصوات اللغوية
}
}
\vspace{-10pt}
\subsection{Politics}
\scriptsize {\foreignlanguage{arabic}{
علم السياسة, ديمقراطية, أنظمة الحكم, حكومة, دولة, نظرية السياسة, الحكومة الفيدرالية, الديمقراطية المباشرة, نظام الانتخابات, دستور, حرية, حقوق الإنسان, التعددية السياسية, حزب سياسي, الأنظمة السياسية, الانتخابات, مؤسسة الرئاسة, نظام الملكية, الاشتراكية, الرأسمالية, الليبرالية, الفاشية, الشيوعية, القومية, الليبرالية الجديدة, السلطة السياسية, العدالة الاجتماعية, حرب الطبقات, الأيديولوجيا السياسية, التخطيط الاقتصادي, التنمية السياسية, الفساد, التسوية السياسية, النظام الانتخابي, الهيئات الدولية, المجتمع المدني, الإصلاح السياسي, التحول الديمقراطي, الحرية السياسية, التربية السياسية, الصراع السياسي, التحليل السياسي, الدور الاجتماعي للإعلام, الاستراتيجية السياسية, المشاركة المدنية, حقوق المرأة في السياسة, العدالة الجنائية, التنظيم السياسي, الحركات السياسية, الإصلاح الدستوري, الثورة, العلاقات الدولية, التحليل الدولي, القانون الدولي, النظام الدولي, النزاع الدولي, التعاون الدولي, السياسة الخارجية, السياسة الداخلية, السياسة الاقتصادية, سياسة التعليم, سياسة الصحة, الحكومات المحلية, التنظيم الإقليمي, الهيئات المنتخبة, الشرطة السياسية, الاحتجاجات السياسية, الاستقلال السياسي, المجتمع الدولي, التحليل الانتخابي, المنظمات الدولية, حروب التحرير, المساعدات الدولية, القوة العسكرية, الأمن القومي, العقوبات الدولية, تسوية النزاعات, حقوق الطفل, إصلاح الشرطة, التحليل الانتخابي, الأسلحة النووية, الدمج السياسي, النهج المعتدل, نظام الديمقراطية التمثيلية, نظام العدالة الجنائية, التسلح, النظام الاقتصادي الدولي, السياسة الطاقة, سياسة البيئة, التحولات الديمقراطية, الصراع الأيديولوجي, السياسة العالمية, نظريات السياسة, مفاهيم سياسية, دولة الرفاه, العدالة المجتمعية, توزيع السلطة, الحقوق السياسية, أزمة الهوية, فكر سياسي, النظام الرياضي السياسي, الديمقراطية الاجتماعية, الحقوق الاقتصادية, السياسة الحزبية, الدولة الأم والطفل, القيادة السياسية, النظام المالي, السياسة الاقتصادية التنموية, التفاعلات السياسية, التصويت, نظرية العدالة, السلطة المستقلة, الحرب التجارية, الحقوق الاجتماعية, البرلمان, سياسة العلاقات الدولية
}
}
\vspace{-10pt}
\subsection{Accounting}
\scriptsize {\foreignlanguage{arabic}{
محاسبة, محاسبة مالية, محاسبة إدارية, قوائم المالية, دفتر الصحيفة, ميزانية عمومية, محاسبة تكاليف, الميزانية, محاسبة الضرائب, محاسبة المراجعة, الرقابة الداخلية, محاسبة الاستثمارات, محاسبة الأصول, محاسبة المخزون, التقارير المالية, التحليل المالي, التكلفة الثابتة, التكلفة المتغيرة, التحليل العمليات, تدفق نقدي, التحليل الإفصاح, المحاسبة المهنية, محاسبة الربح والخسارة, المحاسبة الدولية, محاسبة الشركات, الأعمال المحاسبية, الميزانية التقديرية, تحليل الميزانية, المحاسبة الضريبية, التصنيف المحاسبي, مبادئ المحاسبة, محاسبة المصروفات, محاسبة الإيرادات, مراجعة المالية, التحليل المحاسبي, محاسبة التدفقات النقدية, مراجعة الحسابات, دورة المحاسبة, المحاسبة الاحتياطية, الإفصاح المحاسبي, التسوية المحاسبية, التوازن المالي, التحليل المحاسبي المالي, المحاسبة الصناعية, مقارنة التكاليف, المحاسبة المالية المتقدمة, محاسبة الأعمال المتوسطة, إعداد التقارير المالية, محاسبة التكلفة المعيارية, محاسبة المقاولات, محاسبة الميزانية, التحليل النقدي, مفهوم المحاسبة, مراجعة التقارير المالية, تأثير الضرائب على المحاسبة, مراجعة الحسابات الضريبية, تقارير المحاسبة, الإفصاح عن المالية, تحليل الميزانية العمومية, المالية المحاسبية, البيانات المالية, المحاسبة الإلكترونية, المحاسبة المصرفية, النظام المحاسبي, أثر التكنولوجيا على المحاسبة, النمذجة المحاسبية, المحاسبة التكاملية, محاسبة إعادة التقييم, مراجعة المخزون, مراجعة الميزانية, الفائدة المحاسبية, المحاسبة المالية المتقدمة الضرائب, إدارة التكاليف, التحليل المحاسبي الداخلي, المحاسبة المالية للشركات, المحاسبة الضرائبية, المحاسبة الرقمية, الإيرادات المحاسبية, إدارة المخزون المحاسبي, مراجعة حسابات الضرائب, المحاسبة المقارنة, مراجعة المؤسسات المحاسبية, المحاسبة المصرفية والمالية, المحاسبة المهنية المالية, دورة التحليل المحاسبي, مفهوم التدفقات النقدية, تحليل التكاليف والأرباح, التكلفة المعيارية المحاسبية, أنواع التكلفة المحاسبية, مراجعة الضرائب المحاسبية, محاسبة المدفوعات المالية, مراجعة المالية الشاملة, محاسبة النفقات, نظام الضمانات المحاسبية, دراسة المالية التحليلية, الإدارة المحاسبية, التقارير المالية المتقدمة, إعداد التقارير المحاسبية, أنظمة المحاسبة الدولية, المحاسبة المتقدمة للضرائب, أثر المحاسبة في الاقتصاد, إعداد التقارير المالية الدولية, المحاسبة القطاعية, محاسبة العمليات التجارية, المحاسبة الإنتاجية, التنظيم المحاسبي, مراجعة الملفات المالية, الميزانية الداخلية, تخطيط الماليات المحاسبية, التكلفة الغير مباشرة, محاسبة المصروفات الزمنية.
}
}
\vspace{-10pt}
\subsection{Commerce}
\scriptsize {\foreignlanguage{arabic}{
تجارة, تجارة دولية, تجارة إلكترونية, سوق مالية, اقتصاد تجاري, سوق الأسهم, شركات, صناعة, تسويق, نقل بضائع, شحن دولي, استيراد, تصدير, شركة مساهمة, عقود التجارة, صادرات, دورة السلع, تجارة تجزئة, أسواق مال, بورصة, أسواق الأوراق المالية, إعلانات تجارية, تجارة مباشرة, تجارة الجملة, تجارة مقايضة, قانون التجارة, دفع بالأجل, منشآت تجارية, مجال التجارة, تحليل السوق, اقتصاد التجارة, تجارة محلية, وسائل التجارة, تجارة السلع الأساسية, تجارة السلع المعمرة, تجارة البضائع الجاهزة, تجارة السلع الخدمية, تجارة موسمية, التجارة بين الدول, تجارة الأراضي, اقتصاد تجاري دولي, تجارة القيم, تجارة مستدامة, تجارة محمولة, تجارة عالمية, خدمات تجارية, تجارة مخاطر, تجارة خارجية, تجارة عقارية, تجارة حرة, تجارة مقابلة, التجارة الرقمية, شبكة تجارية, استراتيجية تجارية, صناعة التجارة, خدمات معلومات التجارة, نظام التجارة, أسواق المنتجات التجارية, قيمة التجارة, أسواق التجارة المحلية, أدوات التجارة, نقل تجاري, حركة التجارة, تجارة تجارية, تجارة إلكترونية عالمية, تجارة الأعمال, تجارة العقارات, تجارة منتجات صناعية, تجارة تكنولوجيا المعلومات, تجارة المنتجات العضوية, تجارة الخدمات المالية, تجارة المواد الخام, تجارة التقنيات المتقدمة, نقل المنتجات التجارية, ممارسات التجارة, سياسة التجارة, إدارة التجارة, تجارة الأعمال الدولية, تجارة التكنولوجيا المعمارية, استيراد وتصدير, تجارة المنتجات الزراعية, بروتوكولات التجارة, تجارة الأصناف التجارية, تجارة الخدمات التجارية, تجارة الأعمال الاحترافية, تجارة الكماليات, التجارة الأمامية, التجارة الداخلية, تجارة البضائع الجمركية, تجارة دولية مباشرة, تجارة المنتجات الرقمية, الضرائب التجارية, الشحن التجاري, الحدود التجارية, التجارة الرقمية والإلكترونية, النقل التجاري الدولي, تجارة المواد الغذائية, إدارة سلسلة الإمداد, معايير التجارة, تجارة المنتجات المالية, تجارة مستدامة دولية, التجارة البيئية, أسواق السلع العالمية, شروط التجارة, تحليل التجارة, نمو التجارة, التجارة الآمنة, صناعة التجارة المحلية, الأعمال التجارية الدولية, التجارة الاجتماعية, التجارة العالمية المتقدمة, مراجعة التجارة, تسويق التجارة, تجارة البضائع المستوردة, مراقبة التجارة, أنظمة التجارة, تجارة النفط, التجارة التجارية العالمية, تحليل الأسواق التجارية, الأنشطة التجارية.
}
}
\vspace{-10pt}
\subsection{Industry}
\scriptsize {\foreignlanguage{arabic}{
صناعة, صناعة الحديد, صناعة الصلب, صناعة الأسمنت, صناعة الأدوية, صناعة البتروكيماويات, صناعة الورق, صناعة الأغذية, صناعة المشروبات, صناعة المواد الخام, صناعة الأجهزة الإلكترونية, صناعة السيارات, صناعة الآلات, صناعة الملابس, صناعة الجلود, صناعة الأثاث, صناعة الأحذية, صناعة المعدات الطبية, صناعة الذهب, صناعة المجوهرات, صناعة الزجاج, صناعة الخشب, صناعة الأسلاك, صناعة الأجهزة الكهربائية, صناعة الأدوات المنزلية, صناعة المباني, صناعة البلاستيك, صناعة الطائرات, صناعة القطارات, صناعة الكمبيوتر, صناعة الإلكترونيات الاستعمالية, صناعة الآلات الحاسبة, صناعة الطاقة, صناعة النفط, صناعة الغاز, صناعة الكهرباء, صناعة الفحم, صناعة الصلب الإلكتروني, صناعة الخرسانة, صناعة الحديد المختلط, صناعة الوقود, صناعة المنتجات البتروكيماوية, صناعة المنتجات النفطية, صناعة الزيوت, صناعة المعدات الرياضية, صناعة الإضاءة, صناعة الأجهزة الطبية, صناعة الحقائب, صناعة الأدوات الكتابية, صناعة المنتجات الكيميائية, صناعة البضائع الاستهلاكية, صناعة المواد الكيماوية, صناعة الدواء, صناعة الأدوات الطبية, صناعة الأثاث الخشبي, صناعة الصابون, صناعة المنتجات الخشبية, صناعة الصناعات الخفيفة, صناعة العطور, صناعة الملابس الجاهزة, صناعة التحف, صناعة المنتجات الطبيعية, صناعة الآلات الموسيقية, صناعة الساعات, صناعة المنتجات الورقية, صناعة أدوات المطبخ, صناعة السلع الجمركية, صناعة الملابس الجاهزة, صناعة الجلود المعالجة, صناعة التجهيزات الصناعية, صناعة الدهانات, صناعة الأغطية, صناعة الأثاث المكتبي, صناعة الطب البيولوجي, صناعة المواد العضوية, صناعة المنتجات الزراعية, صناعة البحث والتنمية, صناعة الطائرات المسيرة, صناعة الروبوتات, صناعة الأجهزة الذكية, صناعة المعدات المكتبية, صناعة المنتجات النووية, صناعة الخشب والأثاث, صناعة الفيديو, صناعة الإلكترونيات المتقدمة, صناعة الحواسيب العملاقة, صناعة الطاقة النووية, صناعة الأنظمة الإلكترونية, صناعة المركبات, صناعة الأسلحة, صناعة الألواح الطاقة الشمسية, صناعة الأنظمة الرقمية, صناعة السلع الكتابية, صناعة الملابس المهنية, صناعة مستحضرات التجميل, صناعة المنتجات الطبية, صناعة السلع الاحتياطية, صناعة المعدات الالكترونية, صناعة المعادن, صناعة المواد العضوية الطبيعية, صناعة المنتجات الإلكترونية الاستعمالية, صناعة المنسوجات, صناعة المواد البنائية, صناعة المنتجات المتجددة, صناعة البحث الصناعي, صناعة المنتجات المعدنية, صناعة الطاقة المتجددة, صناعة الورق والمنتجات الورقية, صناعة الكتل الصلبة, صناعة المواد الصناعية.
}
}
\vspace{-10pt}
\subsection{Philosaphy}
\scriptsize {\foreignlanguage{arabic}{
فلسفة, فلسفة العلوم, فلسفة المنطق, فلسفة التاريخ, فلسفة الأخلاق, فلسفة اللغة, فلسفة العقل, فلسفة الوجود, فلسفة الدين, فلسفة الفكر, الفلسفة القديمة, الفلسفة الإسلامية, فلسفة الشرق, الفلسفة الغربية, فلسفة النهضة, فلسفة التنوير, فلسفة التجريبية, الفلسفة المعاصرة, فلسفة المثالية, الفلسفة المادية, فلسفة الوجودية, فلسفة العدالة, فلسفة الحرية, فلسفة الخير والشر, فلسفة الإنسان, فلسفة الجمال, فلسفة الهوية, فلسفة السياسة, فلسفة الحياة, فلسفة المنهج, فلسفة التربية, الفلسفة الوجودية الإسلامية, فلسفة الوعي, فلسفة الحقيقة, فلسفة الفكر الإنساني, فلسفة العقلانية, فلسفة العاطفة, فلسفة المعرفة, فلسفة الحرية والعدالة, فلسفة الكون, فلسفة الديالكتيك, الفلسفة الإستراتيجية, فلسفة العقل البشري, فلسفة الخيال, فلسفة الطبيعة, فلسفة العقل الإنساني, فلسفة الجهاز العصبي, فلسفة الأنطولوجيا, فلسفة المنهجية, فلسفة الوجودية الحديثة, فلسفة التحليل اللغوي, فلسفة الإدراك, فلسفة العدمية, فلسفة التأويل, فلسفة الحداثة, فلسفة التنوير الإسلامي, فلسفة الموضوعية, فلسفة الوجودية الفرنسية, فلسفة الشك, فلسفة الوجودية الألمانية, فلسفة الإمبريقية, فلسفة الواقع, فلسفة الوعي الذاتي, فلسفة الإنسانية, فلسفة الذات, فلسفة الوجود الإسلامية, فلسفة الفكر العلمي, فلسفة المعرفة العملية, فلسفة الموضوعية الإنسانية, فلسفة التساؤل, فلسفة المنطق الفلسفي, فلسفة الحديث, فلسفة العقلانية المعاصرة, فلسفة الديالكتيك الماركسي, فلسفة الواقع الموضوعي, فلسفة الوجودية المعاصرة, فلسفة الفلسفة, فلسفة المنهج العلمي, فلسفة الطبيعة والفكر, فلسفة العقل الإنساني المعاصر, فلسفة الواقع والحقيقة, فلسفة الانتقائية, فلسفة الضوء والظلام, فلسفة الحواس, فلسفة الذاتية, فلسفة الأنطولوجيا المعاصرة, فلسفة الحتمية, فلسفة الروح, فلسفة الوجود المعاصر, فلسفة التسوية, فلسفة الوعي الجماعي, فلسفة العدمية المعاصرة, فلسفة الإنسان المعاصر, فلسفة الحس المشترك, فلسفة الوجودية المتأخرة, فلسفة الوجود الإنساني, فلسفة الذكاء الاصطناعي, فلسفة الاستقلالية, فلسفة الزمن, فلسفة الحوار, فلسفة الاستقلال, فلسفة الذات والعقل, فلسفة التجريد, فلسفة الوظيفة, فلسفة الروابط البشرية, فلسفة العقل الجماعي, فلسفة العقل المعنوي, فلسفة الظواهر, فلسفة الاختلاف, فلسفة الاختيار, فلسفة المعرفة الذاتية, فلسفة التفاعل, فلسفة العلاقات الإنسانية, فلسفة الروح المعاصرة, فلسفة الإنسان الحديث, فلسفة المعرفة العقلية.
}
}
\vspace{-10pt}

\subsection{War}
\scriptsize {\foreignlanguage{arabic}{
حرب, حروب العالم, حرب العالم الأولى, حرب العالم الثانية, حرب الولايات المتحدة الأهلية, الحرب الباردة, حرب فيتنام, الحرب العراقية الإيرانية, حرب الخليج الثانية, حرب فلسطين 1948, حرب الاستقلال الإسرائيلية, حرب التحرير الجزائرية, حرب العدوان الست يوم, حرب الكرمل, حرب التحرير الأفغانية, حرب التحرير اللبنانية, حرب الاستنزاف, حرب العصابات, حرب الطائفيين, الحرب الاقتصادية, الحرب الإلكترونية, حرب المستقبل, الحروب الدينية, الحروب العالمية, الحروب القديمة, الحروب العسكرية, حرب الشبكات, الحرب الإعلامية, حروب الفتح الإسلامي, حرب الولايات المتحدة والمكسيك, حرب اليابان الصينية الأولى, حرب اليابان الصينية الثانية, حرب العالم الأولى في الشرق الأوسط, الحرب المجمدة, الحرب الاقتصادية الحالية, الحروب العربية الإسرائيلية, الحرب الآسيوية, حرب الأقاليم, حرب الأمريكتين, الحرب الصليبية, الحروب الأهلية الأمريكية, الحروب الأنغلو فرنسية, حرب الحرير, حروب الإمبراطورية الرومانية, حروب الإمبراطورية العثمانية, حرب الأقاليم الإمبراطورية, حروب القرون الوسطى, حروب المماليك, حرب الاستقلال الولايات المتحدة, الحروب الإنجليزية الإسبانية, حروب التوسعات الإسلامية, حروب الفتوحات الإسلامية, حرب الفرنجة, حروب الهجوم الإسلامي, حروب الثوار, حروب اللاجئين, الحروب الحاسوبية, حروب القرصنة, حروب العصور الوسطى, حروب إفريقيا, حروب الاستعمار الأوروبي, حروب التكنولوجيا العسكرية, الحروب الداخلية, حرب التكنولوجيا النووية, حروب الشعب, حروب الجزيرة العربية, حرب الاقتصاد, الحروب الطائفية, حروب المدن, حروب الجيش الحديث, حروب المجاهدين, حروب المستقبل العمليات, حروب الفضاء, حرب النفط, الحروب العالمية الثالثة, حروب الملوك, الحروب الاقتصادية في القرن 21, حروب الثورات, حرب الفرنجة في الشرق, حروب الحروب, حروب العدوان العسكري, حروب الفتح, حروب الصراع السياسي, الحروب الثقافية, حروب الأنظمة, حروب الطبقات الاجتماعية, حروب الاستقلال, حروب التقسيم, حروب التدمير, حروب المؤامرات, حروب الشرق الأوسط, حروب النظام العالمي, حروب الانتشار الإسلامي, حروب التطهير الإثني, حروب الفقر, حروب القتال المباشر, حروب المنطقة العربية, حروب اللاجئين المسلحين, حروب الاستراتيجية, حروب القتال البحري, حروب القتال البري, حروب القتال الجوي, حروب المناورات القتالية, حروب القتال في المناطق المدنية, حروب الحضارة, حروب العدوان, حروب الإغارة, حروب التهجير الجماعي, حروب الاشتباك, حروب التجربة العسكرية, حروب الصراعات الداخلية, حروب المساعدات الإنسانية, حروب القتال الدولي, حروب العنف المسلح, حروب التصعيد, حروب التخويف النووي, حروب الاندماج, حروب المدافع, حروب السلطة, حروب التسلح الدولي, حروب المناطق الساخنة, حروب الساعة صفر
}
}

\end{document}